%% file: paper.tex
\documentclass[letterpaper]{article} 
\input{commands}

\usepackage[disable]{todonotes}

\title{Implicit State and Goals in QBF Encodings for Positional Games \\ (extended version)}
\author {
    Irfansha Shaik\textsuperscript{\rm 1} \orcidID{0000-0002-7404-348X},
    Valentin Mayer-Eichberger\textsuperscript{\rm 2},
    Jaco van de Pol\textsuperscript{\rm 1}  \orcidID{0000-0003-4305-0625},
    Abdallah Saffidine\textsuperscript{\rm 3}
}
\affiliations {
    \textsuperscript{\rm 1} Aarhus University, Department of Computer Science, Aarhus, Denmark\\
    \textsuperscript{\rm 2} University of Potsdam, Germany, valentin@mayer-eichberger.de\\
    \textsuperscript{\rm 3} The University of New South Wales, Sydney, Australia, abdallahs@cse.unsw.edu.au\\
}

\usepackage{amsmath}
\usepackage{amssymb}
\usepackage{booktabs,multirow}

\newcommand{\jaco}[1]{\todo[inline,color=blue!50]{\textbf{JvdP:} #1}}

\usepackage{mathtools}

\usepackage{amsthm}
\newtheorem{definition}{Definition}

\DeclareMathOperator{\move}{M} 
\DeclareMathOperator{\nodes}{W} 
\DeclareMathOperator{\allnodes}{\mathcal{P}} 
\DeclareMathOperator{\sympos}{P} 
\DeclareMathOperator{\state}{S} 
\DeclareMathOperator{\I}{I} 
\DeclareMathOperator{\Goal}{G} 

\DeclareMathOperator{\T}{T} 
\DeclareMathOperator{\neighbor}{N} 
\DeclareMathOperator{\bound}{BR} 

\DeclareMathOperator{\n}{\eta} 
\DeclareMathOperator{\pathlength}{\ell} 

\DeclareMathOperator{\W}{w} 
\DeclareMathOperator{\B}{b} 

\DeclareMathOperator{\bin}{bin} 

\DeclareMathOperator{\border}{\Gamma} 
\newcommand{\borders}{\border_s} 
\newcommand{\bordere}{\border_e} 

\DeclareMathOperator{\white}{w} 
\DeclareMathOperator{\occupied}{o} 

\DeclareMathOperator{\win}{win} 
\DeclareMathOperator{\WIN}{\mathcal{W}} 
\DeclareMathOperator{\EDGES}{\mathcal{E}} 

\DeclareMathOperator{\src}{SC} 
\DeclareMathOperator{\trg}{TG} 

\DeclareMathOperator{\UPG}{\Pi} 
\DeclareMathOperator{\play}{\phi} 

\DeclarePairedDelimiter\ceil{\lceil}{\rceil}

\newcommand{\bblack}[2]{\mathsf{b}_{#1}^{#2}}
\newcommand{\bwhite}[2]{\mathsf{w}_{#1}^{#2}}
\newcommand{\bmove}[2]{\mathsf{m}_{#1}^{#2}}
\newcommand{\bhedge}[1]{\mathsf{h}_{#1}}
\newcommand{\breach}[1]{\mathsf{r}_{#1}}
\newcommand{\bpath}[2]{\mathsf{p}_{#1}^{#2}}

\newcommand{\smove}[1]{\mathsf{M}^{#1}}
\newcommand{\sposi}[1]{\mathsf{P}^{#1}}

\newcommand{\binaryp}[1]{\overline{#1}}
\newcommand{\binaryn}[1]{\underline{#1}}
\newcommand{\nbnodes}{\n}

\newcommand{\size}[1]{\,|\!#1\!|\,}

\begin{document}

\maketitle              
\begin{abstract}
We address two bottlenecks for concise QBF encodings
of maker-breaker positional games, like Hex and Tic-Tac-Toe. Our baseline
is a QBF encoding with explicit variables for board positions and an explicit
representation of winning configurations. 
The first improvement is inspired by
lifted planning and avoids variables for explicit board positions, 
introducing a universal quantifier representing a symbolic board state.
%
The second improvement represents the
winning configurations implicitly, exploiting their structure. 
%
The paper evaluates the size of several encodings, depending on board size and game depth. 
It also reports the performance of QBF solvers on these encodings.
We evaluate the techniques on
Hex instances and also apply them to Harary's Tic-Tac-Toe.
In particular,
we study scalability to 19$\times$19 boards, played in human Hex tournaments.
\end{abstract}

\section{Introduction}
\label{sec:introduction}

This paper presents new encodings of positional games in Quantified Boolean Logic (QBF).
In these games, two players claim empty positions on a fixed board in alternating turns.
Examples include Hex, Harary's Tic-Tac-Toe (HTTT), and Gomoku.
In the maker-breaker variant, the first player wins if he manages to occupy some
winning configuration, while the second player wins if she can avoid that.

Quantifier alternations in QBF can naturally express the existence of a winning strategy of bounded depth.
This allows solving these PSPACE-complete games using the sophisticated search techniques of generic QBF solvers.
The quality (size and structure) of the encoding has a great influence on performance,
but precise knowledge of what constitutes a good encoding is quite limited.

We address two bottlenecks in current QBF encodings for positional games~\cite{10.1007/978-3-030-51825-7_31}: First, unrolling to bounded depth
causes duplication of variables and clauses. Extending techniques from planning~\cite{DBLP:conf/aips/ShaikP22},
we represent symbolic positions by universal variables. This lifted encoding is concise, at the expense of
an extra quantifier alternation. We also study stateless encodings, which avoid position variables
by expressing all constraints in terms of the chosen moves.

The other bottleneck is the representation of winning configurations. In some positional games,
the winning configurations consist of a fixed number of shapes (Tic-Tac-Toe, Gomoku). However,
in Hex, winning configurations are defined in terms of paths that connect two borders of
the board. So an explicit representation of the goal is exponential in the
board size. We study implicit goal representations, by expressing winning path conditions using 
neighbor relations.
It appears that implicit goal constraints not only yield more concise encodings but also boost
the performance of current QBF solvers by an order of magnitude. The fastest result was obtained
by an implicit encoding of the winning condition in the \emph{transversal} game (the second player cannot win).

In this paper, we propose the following 7 QBF encodings, combining all ideas mentioned above.
We study the size of these encodings depending on board size and game depth.
The encoded games can be solved with existing QBF solvers.
We present an experimental evaluation, measuring the performance of a QBF solver for all our encodings,
applied to a benchmark of Piet Hein's Hex puzzles (on 3$\times$3 - 7$\times$7 boards),
to a set of human-played Hex championship plays (19$\times$19 board), and to Harary's Tic-Tac-Toe (5$\times$5).

\smallskip
\begin{center}
\noindent
\begin{tabular}{@{}l | lll@{}}
\hline
Goal: ~~~/~~~ Board:                      & Explicit & Lifted & Stateless \\
\hline
All minimal paths                  & \ref{subsubsec:hyperedge-goal-hex} (EA) & \ref{subsec:explicitgoalsym} (LA)           & \ref{subsec:witnessbasedexplicit} (SA)        \\
Neighbor-based            & \ref{subsec:expb-level-goal-hex} (EN)       & \ref{subsec:symblevelnum} (LN)         & \ref{subsec:statelesslevelnum} (SN)     \\
Transversal game           & \ref{subsec:transversal-goal-hex} (ET) & -           & -       \\
\hline
\end{tabular}
\end{center}

\section{Preliminaries}
\subsection{Maker-Breaker positional games and QBF}
\label{sec:qbfandpositionalgames}

A positional game is played on a board, where 2 players, called Black and White, occupy empty positions in turns.
The initial board can be either empty or some of the positions are already occupied.
In the maker-breaker variant, a game is won by the first player (Black) if and only if 
the final set of black positions contains a winning set;
otherwise, it is won by the second player (White), so there is no draw in these games.
We define these games formally as follows.
\begin{definition}
\label{def:makerbreakergame}
Given a set of positions $\allnodes$, define $\n=\size{\allnodes}$.
A maker-breaker positional game $\UPG$ is a tuple $\langle \I, \WIN\rangle$, with
\begin{itemize}
\item Initial state $\I = (\I_{\B}, \I_{\W})$ s.t.\ $\I_{\B}, \I_{\W} \subseteq \allnodes$ and $\I_{\B} \cap \I_{\W} = \emptyset$,
\item and Goal condition $\WIN \subseteq 2^{\allnodes}$.
\end{itemize}
\end{definition}
We assume, without loss of generality, that $\I_{\B}=\I_{\W}=\emptyset$, possibly after some preprocessing (cf.\ Section~\ref{subsec:gex}).

A single play is a sequence of moves (occupying positions) chosen by players in turns.
We assume Black plays first and in a maximal play, the game ends with Black's turn.

\begin{definition}
\label{def:play}
Given $\UPG$, a single play $\play$ is a sequence of $k$ positions $\langle\play_{1}, \dots, \play_{k}\rangle$ chosen by each player alternatively. The black moves $\play_{\B} = \{\play_{i} \mid 1\leq i \leq k, i \text{ odd}\}$ and the white moves $\play_{\W} = \{\play_{i} \mid 1\leq i \leq k, i \text{ even}\}$.
A play is valid when $\{\play_{\B} \cup \play_{\W}\}  \cap \{\I_{\B} \cup \I_{\W}\} = \emptyset$ and $\play_{i} \neq \play_{j}$ for all $i \neq j$.
\end{definition}

\begin{definition}
\label{def:winningcondition}
Given $\UPG$ and a valid play $\play$, we say the play $\play$ is won by Black if and only if there exists a set of positions $\win \in \WIN$ such that $\win \subseteq \I_{\B} \cup \play_{\B}$.
\end{definition}



The goal of all encodings in this paper is to decide if Black has a winning strategy, i.e.,
Black will win all valid games where it plays according to this strategy.
We consider only strategies for a bounded number of moves up to depth $d$.



\subsection{Hex and Generalized Hex}
Hex is a well-known positional game played on an $n\times n$-board of hexagons, such that each (non-border) position has six neighbors~\citep{HaywardT2019}.
The game is won by Black if there is a black path connecting Black's two opposite borders.
On a completely filled board, Black has a winning connection if and only if White's opposite borders are not connected with a path of white stones.
This is known as the \emph{Hex Theorem}~\citep{Gale1979}.

Because of its simplicity and rich mathematical structure, Hex has provided a source of inspiration for the design and implementation of specialized solvers~\citep{arneson2009mohex},
as well as for theoretical considerations on computational complexity~\citep{Reisch1981,BonnetJS2016,BonnetGLRS2017}.
\begin{definition}
\label{def:hexgame}
\emph{Generalized Hex} is a 2-player game between Short and Cut, where an instance is a $3$-tuple $\langle G, s, e \rangle$, with
$G = \langle \allnodes\cup\{s,e\}, \EDGES\rangle$ an undirected graph, and $s$ and $e$ two distinguished nodes.
The two players take turns claiming nodes from $\allnodes$ and Short wins if $s$ to $e$ are connected with a path of nodes claimed by Short.
Cut aims at preventing it by claiming a set of nodes that constitutes a cut from $s$ to $e$.
\end{definition}
To simplify notation in our encodings, we introduce $\border_{s} = \{ v \mid (s, v) \in \EDGES \}$ and $\border_{e} = \{v \mid (v,e) \in \EDGES\}$.
Short's goal is to create a path from any node of $\border_{s}$ to any node of $\border_{e}$.

\subsection{Quantified Boolean Formulas}
We consider closed QBF formulas in prenex normal form, i.e., of the form $Q_1 x_1 \cdots Q_n x_n (\Phi)$, where
$\Phi$ is a propositional formula with Boolean variables in $\{x_1,\ldots,x_n\}$ and each $Q_i\in\{\forall,\exists\}$. 
Every such formula evaluates to true or false. QBF evaluation is a standard PSPACE-complete problem. It is well
known that the complexity increases with the number of quantifier alternations.

Several QBF solvers exist, which operate on QBF in either QDIMACS format, where $\Phi$ is essentially a set of CNF
clauses, or in QCIR format \citep{jordan2016noncnf}, where $\Phi$ is provided as a circuit with and- and or-gates and negation.
In general, QDIMACS is more low-level and allows efficient operations, while QCIR contains more
structure and can be more readable.
We use both QDIMACS, using Bule \citep{JungMES2022} for concise specifications, and QCIR, which can be 
transformed to QDIMACS using the \citet{Tseitin1983CDPC} transformation, 
introducing one existential Boolean variable per gate.

\section{Maker-breaker explicit goal encodings}
\label{sec:expgoalqbfencoding}

All our encodings provide a QBF formula that evaluates to True if and only if the corresponding game is
won by Black from the empty board, using $d$ moves starting and ending with Black (so $d$ is odd).
Using QBF, we can naturally capture the moves of the players in a maker-breaker positional game 
by $d$ alternating existential and universal variables.


\subsection{Explicit board (EA)}
\label{subsec:expboardexpgoal}
Our baseline optimizes the corrective encoding COR by \citet{10.1007/978-3-030-51825-7_31}.
The main idea is to unroll the transition relation $d$ times. 
In COR, board positions were maintained after each move. In EA, we save many frame conditions
by maintaining board positions after black moves only. These are needed to test the validity 
of the moves and the winning condition. All moves are encoded logarithmically (in COR only the
White moves). 

We identify the set of positions $\allnodes$ with the integers $\{0, 1, \dots, \nbnodes-1\}$.
For any $v \in \allnodes$, we consider the binary representation of $v$ over $\ceil{\lg \nbnodes}$ bits.
We write $\binaryp{v}$ for the set of bits assigned 1 in $v$ and $\binaryn{v}$ for the set of bits assigned 0 in $v$.
For instance, if $v = 5 = b00\dots101$, we have $\binaryp{v} = \{0, 2\}$ and $\binaryn{v} = \{1, 3, 4, \dots, \ceil{\lg\nbnodes}-1\}$.

\subsubsection{Explicit board maintenance.}
\label{subsubsec:expboard}
We introduce alternating variables $\smove{t}$ for the chosen move at step $t$ and $\sposi{t}$ to represent the board at odd time steps $t$ (after each black move).
\begin{align}
& \exists \smove{1} \sposi{1} \forall \smove{2} \exists \smove{3} \sposi{3} \dots \forall \smove{t-1} \exists \smove{t} \sposi{t} \dots \exists \smove{d} \sposi{d} & \label{eq:1}
\end{align}
Here variables $\smove{t} = \{ \bmove{i}{t} \mid 0 \leq i < \ceil{\lg \nbnodes} \}$ and variables $\sposi{t} = \{ \bblack{v}{t}, \bwhite{v}{t} \mid v \in \allnodes \}$.

For each vertex $v \in \allnodes$, odd time step $t$, and bit index $j$ we get the following clauses, which specify
the correct color of each position at odd time steps $t$, depending on the previous position and the chosen move.
\begin{align}
\bwhite{v}{t-2} &\rightarrow \bwhite{v}{t} & \\
\bigwedge_{i \in \binaryp{v}} \bmove{i}{t-1} \wedge \bigwedge_{i \in \binaryn{v}} \neg \bmove{i}{t-1} \wedge \neg \bblack{v}{t-2} &\rightarrow \bwhite{v}{t} &\\
\bwhite{v}{t} &\rightarrow \neg \bblack{v}{t} & \\
\phantom{\neg} \bmove{j}{t} \wedge \neg \bblack{v}{t-2} &\rightarrow \neg\bblack{v}{t} & \text{if }j\in \binaryn{v}\\
\neg \bmove{j}{t} \wedge \neg \bblack{v}{t-2} &\rightarrow \neg\bblack{v}{t} & \text{if }j\in \binaryp{v}\label{eq:6}
\end{align}
When the above clauses refer to variables unquantified in the prefix, e.g., $\bblack{v}{-1}$, they are substituted with $\bot$.
Note that when $\bblack{v}{t}$ is not forced to False by the clauses above, it becomes True by the existential quantification over $\sposi{t}$.

\subsubsection{Hyperedge-based goal detection.}
\label{subsubsec:hyperedge-goal-hex}

We introduce variables $\bhedge{h}$ in the innermost existential block, indicating that Black won by configuration $h\in\WIN$.
\begin{align}
& \exists \{\bhedge{h} \mid h \in \WIN \} & \label{eq:7}
\end{align}

The following clauses indicate that all positions in some winning configuration 
are black in the final state, i.e., after $d$ moves:
\begin{align}
& \bigvee_{h \in \WIN} \bhedge{h} & \\
& \bhedge{h} \rightarrow \bblack{v}{d} & \text{for } h \in \WIN\text{, and } v \in h\label{eq:9}
\end{align}

\subsection{Lifted Board (LA)}
\label{subsec:explicitgoalsym}
The previous encoding generates similar variables and constraints for each position after a black move.
Note that the validity of all moves can be checked independently for each individual position.
We now use this structure to generate a lifted encoding, representing position constraints symbolically on a universal position variable.
This follows the lifted encoding for planning in \citet{DBLP:conf/aips/ShaikP22}.

We introduce variables $\move = \{\move^{1}, \dots, \move^{d}\}$,
where $\move^{t}$ is a sequence of $\ceil{\lg(\n)}$ boolean variables representing one move at time step $t$.
Given a single play, we check if all moves are valid by using a single symbolic position,
represented by $\sympos$, a sequence of $\ceil{\lg(\n)}$ universal variables.
When expanded, the universal branches of $\sympos$ correspond to all board positions.
We use 2 variables for each time step to represent the state of the symbolic position.
Let $\state^t = \{\occupied^{t}, \white^{t}\}$, representing if the symbolic state is \emph{occupied} and/or \emph{white}
at time step $t$.
We introduce witness variables $\nodes = \{\nodes^{0}, \dots, \nodes^{\pathlength}\}$, 
i.e., a sequence of positions that should form a winning configuration for black.
Here $\pathlength=\frac{d-1}{2}$, so $\pathlength+1$ is the maximum size of a witness set in $\WIN$.
Each $\nodes^{i}$ is a sequence of $\ceil{\lg(\n)}$ boolean variables.

The corresponding Lifted Encoding for finding a winning strategy for Black with explicit goals has this shape:
\begin{align*}
&\exists \move^1 \forall \move^2 \dots \exists \move^{d}\\
&\exists \nodes^{0}, \dots, \nodes^{\pathlength} \\
&\forall \sympos \\
&\exists \state^1, \dots, \state^{d+1} \\
&\Goal_{\win}(\nodes) \land \Goal_{sb}(\state^d, \nodes, \sympos) \land \neg\occupied^{1} \land {}\\
&\bigwedge_{t=1,3, \dots, d}\T^{t}_{\B}(\state^t, \state^{t+1}, \sympos, \move^t) \land \bound(\move) \land {}\\
&\bigwedge_{t=2,4, \dots, d-1}\T^{t}_{\W}(\state^t, \state^{t+1}, \sympos, \move^t)
\end{align*}

The initial board is empty, hence $\neg\occupied^{1}$.
This forces \emph{every} node to be unoccupied, since it is under the universal quantifier $\forall\sympos$.
There are 2 goal constraints: 
(1) the witness must be an element of $\WIN$ ($\Goal_{\win})$;
(2) witness positions must be black ($\Goal_{sb}$), i.e., the position at time step $d$ in the corresponding
universal branch must be occupied and not white.

We use ``$\bin$'' to express the logarithmic encoding of positions on the board by $\ceil{\lg(\n)}$ bits.
That is, $\bin(x_{\ceil{\lg(\n)}-1}\dots x_1x_0,v) = {\bigwedge_{i \in \binaryp{v}} x_{i}} \wedge {\bigwedge_{i \in \binaryn{v}} \neg x_{i}}$.
We also use ``='' to denote the bit-wise conjunction of bi-implications.

\begin{definition}
\label{def:goalconstraint}
Explicit goal constraints:
\begin{align*}
\Goal_{sb}(\state^d, \nodes, \sympos) =
\big(\bigvee_{i=0}^{\pathlength} (\sympos = \nodes^{i})\big) \implies (\neg \white^{d+1} \land \occupied^{d+1}) \\
\Goal_{\win}(\nodes) = \bigvee_{\{v_{0}, \dots, v_{m-1}\} \in \WIN} \bigwedge_{i=0}^{m-1} \bin(\nodes^{i},v_{i})
\end{align*}
\end{definition}

We separate the transition constraints for Black and White.
For Black, the position in the universal branch of the chosen move must be empty at time step $t$
and black at time step $t+1$. In all other branches, we just propagate the position.
For White, the position is only updated if it is empty in the universal branch of the chosen move. 
Otherwise, the position is propagated.

\begin{definition}
\label{def:blacktransitionfunction}
Black transitions: $\T^{t}_{\B}(\state^t, \state^{t+1}, \sympos, \move^t) =$
\begin{align*}
\big((\sympos = \move^{t}) \implies (\neg\occupied^{t} \land \neg\white^{t + 1} \land \occupied^{t+1})\big) \land {} \\
\big((\sympos \neq \move^{t}) \implies (\white^{t} = \white^{t+1} \land \occupied^{t} = \occupied^{t+1})\big)
\end{align*}
Bound Restriction, black: $\bound(\move) = \bigwedge_{t=1,3,\ldots,d} (\move^{t}\! {<} \n)$
\end{definition}

\begin{definition}
\label{def:whitetransitionfunction}
White transitions: $\T^{t}_{\W}(\state^t, \state^{t+1}, \sympos, \move^t) =$
\begin{align*}
\big((\sympos = \move^{t} \land \neg\occupied^{t} ) \implies ( \white^{t + 1} \land \occupied^{t+1})\big) \land {} \\
\big((\sympos \neq \move^{t} \lor \occupied^{t}) \implies (\white^{t} = \white^{t+1} \land \occupied^{t} = \occupied^{t+1})\big)
\end{align*}
\end{definition}

Finally, the restriction on black moves ($\bound$, Def.~\ref{def:blacktransitionfunction}) limits Black's legal moves
to the board, in case the number of positions is not a power of 2. We use the standard less-than ($<$) on
bit vectors. This prunes the search space. A similar constraint for White reduced the performance of the QBF solver.
It is safe to drop it since White can never win by playing outside the board (it corresponds to giving up a move).
\subsection{Stateless (SA)}
\label{subsec:witnessbasedexplicit}

We now present an encoding that avoids all intermediate states entirely. 
The idea is similar to Causal Planning \cite{DBLP:conf/kr/KautzMS96}.
In a positional game, 
the validity of Black's moves can be ensured by checking that they are different from all previous moves. 
Also, the winning condition can be expressed completely in terms of Black's moves.
In the maker-breaker case, we don't even need to check the validity of White's moves.
We just ignore White's moves that are played outside the board or on occupied positions.
As in~Sec.~\ref{subsec:explicitgoalsym}, this is correct because White cannot win by giving up a move (in a positional game).

The move and witness variables are the same as in Section~\ref{subsec:explicitgoalsym}.
We can now drop all other variables for the SA encoding.
There are three main constraints: 
  (1) the witness is one of the winning sets of positions ($\Goal_{\win}$, cf.~Def.~\ref{def:goalconstraint});
  (2) the witness positions form a subset of Black's moves ($\Goal_{wb}$); 
   (3) Black's moves cannot overwrite previous moves.
\begin{align*}
&\exists \move^1 \forall \move^2 \dots \exists \move^{d}\\
&\exists \nodes^{0}, \dots, \nodes^{\pathlength} \\
&\Goal_{\win}(\nodes) \land \Goal_{wb}(\move, \nodes) \land \bound(\move) \land  {}\\
&\bigwedge_{t=1,3,\dots,d} \bigwedge_{i<t} \move^t \neq \move^i
\end{align*}

Instead of the symbolic black constraint, here we use the black moves to constrain the witness to be black.

\begin{definition}
\label{def:wbgoalconstraint}
Witness black constraint $\Goal_{wb}(\move, \nodes) =$
\begin{align*}
\bigwedge_{i=0}^{\pathlength} \bigvee_{t=1,3,\dots,d} \nodes^i = \move^t
\end{align*}
\end{definition}

Note that the explicit winning condition is the same as the LA encoding.
In all explicit goal encodings, the size grows linearly with the number of explicit winning configurations.

\section{QBF encodings with implicit goal}
\label{sec:implicitgoal}

For Gomoku and Tic-Tac-Toe, the explicit goal representation is good enough,
since the number of winning configurations is polynomial in the board size.
However, in Hex, the winning condition is based on paths, 
whose number grows exponentially in the board size.
Thus, the size of all explicit goal encodings of Section~\ref{sec:expgoalqbfencoding} increases exponentially.

In this section, we represent path conditions compactly, using the structure of the board.
We consider board games as graph games, where the edges indicate the neighbor relation.
This paves the way for compact encodings of paths (and other structures).
We illustrate implicit goal encodings for the Maker-Breaker positional game Hex,
but it is also applicable to other positional games, like HTTT (Sec.~\ref{sec:implicitgttt}).
Section~\ref{subsec:gex} discusses preprocessing steps from Hex to Generalized Hex.
The subsequent subsections apply compact goal encodings to Explicit, Lifted, and Stateless boards.
The last subsection considers the transversal game, avoiding White's connecting paths instead.

\subsection{Transformations to Generalized Hex}
\label{subsec:gex}
Consider the Hex puzzle in Figure~\ref{fig:9-original} due to Hein~\citep{HaywardT2019}.
Black has a winning strategy of depth $7$ starting with $c_3$.
For the sake of a running example, we will assume we search for a strategy of depth $d=5$.
Let us call $B$ the existence of a $5$-move win for Black on Figure~\ref{fig:9-original}.

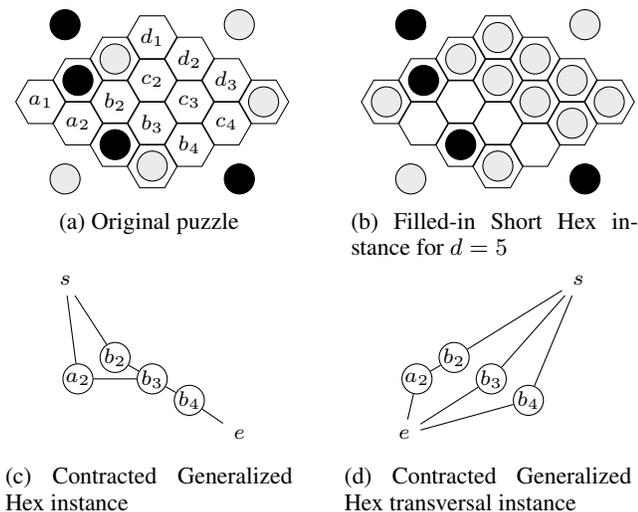
\begin{figure}[t]
\centering
\input{hein9}
\caption{Hein puzzle $9$ and its reductions to Generalized Hex when $d=5$.
Black wins in 5 in (\subref{fig:9-original}) iff Black wins in 5 in (\subref{fig:9-short-hex}) iff Short wins in 5 in (\subref{fig:9-gex}) iff Cut wins in 5 in (\subref{fig:9-transversal-gex}).}
\label{fig:hexexample}
\end{figure}

Recall that any winning strategy of $d$-moves can involve $\ell+1=\frac{d+1}{2}$ black moves at most.
As such, black cannot win through any winning configuration of size $>\ell+1$, so these configurations can be removed from the game without altering whether a $d$-move winning strategy exists.
Similarly, we could remove from $\WIN$ any winning configuration that is not minimal for the subset relation. 
Removing these configurations from $\WIN$ may result in many positions not occurring in any winning configuration at all.
In a Maker-Breaker game, such positions may as well be claimed by White (Breaker).
As such, our initial query, $B$, is equivalent to whether Black has a $5$-move win in Fig.~\ref{fig:9-short-hex}.

In any Maker-Breaker game, positions already claimed by either player can be preprocessed.
Any winning configuration containing a white-claimed position is removed from $\WIN$ and black-claimed positions can be removed from any winning configuration.
Following the contraction of \citet[proof of Theorem 2]{BonnetGLRS2017} for Generalized Hex, this corresponds to removing any Cut-claimed vertex and its incident edges and removing any Short-claimed vertex and turning its neighborhood into a clique.
Applying this contraction to our running example gives Fig.~\ref{fig:9-gex} where Short has a $5$-move win if and only if $B$.

The Hex Theorem provides another approach to solving Hex, which involves the transversal game.
To verify if Black has won after $d$ moves, one can fill the remaining empty cells for White and ensure that even then White has no connecting paths.
For this approach, we can similarly fill unnecessary cells for White (Fig.~\ref{fig:9-short-hex}) and apply the contraction process to Generalized Hex with the two players swapped as in Fig.~\ref{fig:9-transversal-gex}.
This position admits a $5$-move win for Cut iff $B$.

\subsection{Explicit Board Neighbor-based (EN)}
\label{subsec:expb-level-goal-hex}

We first apply the symbolic neighbor-based goal encoding to the explicit board encoding
from Section~\ref{subsec:expboardexpgoal}. To this end, we keep the quantification
and transition constraints (Eq.~\ref{eq:1}-\ref{eq:6}). We replace the explicit
goal constraints (\ref{eq:7}-\ref{eq:9}) by the following constraints.
Recall that $\ell = \frac{d-1}{2}$ is the maximum length of a winning path.
We introduce boolean variables:

\begin{align}
& \exists \{\bpath{v}{i} \mid v \in \allnodes, 0 \leq i \leq \ell \} &
\end{align}
Here $\bpath{v}{i}$ codes that position $v$ is the $i$-th position in the witness path.
We get the following clauses, which specify that all positions on the path are black on the final board;
the initial path position is on the start border, the path is connected in the graph;
and the last path position is on the end border.
\begin{align}
& \bpath{v}{i} \rightarrow \bblack{v}{d} & \text{for } v \in \allnodes, 0 \leq i \leq \ell\\
& \bigvee_{v\in \borders} \bpath{v}{0}\\
& \bpath{v}{i} \rightarrow \bigvee_{(v,w) \in \EDGES} \bpath{w}{i+1} & v \in \allnodes \setminus \bordere, 0 \leq i < \ell \label{eq:en-avoid-stutter}\\
& \neg \bpath{v}{\ell} & v \in\allnodes\setminus\bordere
\end{align}

\subsection{Lifted Neighbor-based (LN)}
\label{subsec:symblevelnum}

We now modify the LA encoding (Sec.~\ref{subsec:explicitgoalsym}) with neighbor-based symbolic goals.
The only change is in the goal constraint, $\Goal_{\win}$.
Instead of explicit winning configurations, we now use the neighbor relation from the Generalized Hex input.
First, we specify that the first and last witness positions lie on the proper borders $(\src \land \trg)$.
We must also specify that adjacent positions in the witness path are connected in the graph. 
To do this in the lifted encoding, we introduce neighbor variables $\neighbor$, a sequence of $\ceil{\lg(\n)}$ variables, 
coding the symbolic neighbor of the symbolic universal position $\sympos$. 
Now we can simply specify that the binary decodings of $\neighbor$ and  $\sympos$ are related in the graph 
(first implication below). Finally, we specify that if some witness position matches $\sympos$ in the current branch,
the next witness position matches its symbolic neighbor $\neighbor$ (second implication below).
So, we replace $\Goal_{\win}(\nodes)$ in Sec.~\ref{subsec:explicitgoalsym} by the following constraints:

\begin{align*}
&\exists \neighbor { } \src \land \trg \land {}\\
&\bigwedge_{v=0}^{\n-1} \big(\bin(\sympos,v) \implies \bigvee_{(v,w) \in \EDGES}\bin(\neighbor,w)\big) \land {}\\
&\bigwedge_{i=0}^{\pathlength-1}{\big((\nodes^{i} = \sympos) \implies (\nodes^{i+1} = \neighbor)\big)}
\end{align*}

\begin{definition}
\label{def:sourceandtargetconstraints}
Source $\src = \bigvee_{v \in \border_{s}}  \bin(\nodes^{0},v)$ and
Target $\trg = \bigvee_{v \in \border_{e}} \bin(\nodes^{\pathlength},v)$
\end{definition}

There is a slight complication: the winning paths can be of different lengths, while our encoding assumes that
all paths have the same length $\pathlength+1$. This is solved by allowing some ``stutter steps'' at
the end of the witness path. Rather than changing the QBF encoding, we implemented this by adding the
reflexive closure for $\border_{e}$ nodes to $\EDGES$ in the input graph.

\subsection{Stateless Neighbor-based (SN)}
\label{subsec:statelesslevelnum}

We now show how to avoid the explicit goal constraints in the stateless encoding.
We only replace the winning goal constraint $\Goal_{\win}$ in the SA encoding of Sec.~\ref{subsec:witnessbasedexplicit}.
First, we provide constraints for first and last witness positions, as in the LN encoding (Def.~\ref{def:sourceandtargetconstraints}).
Then, since the stateless encoding has no notion of a symbolic node, we need to provide the neighbor constraints for each pair of adjacent witness nodes.

\begin{align*}
& \src \land \trg \land \\
&\bigwedge_{i=0}^{\pathlength-1} \bigwedge_{v=0}^{\n-1} \big(\bin(\nodes^i,v) \implies \bigvee_{(v,w) \in \EDGES} \bin(\nodes^{i+1},w)\big)
\end{align*}


\subsection{Explicit-Board Transversal-based goal (ET)}
\label{subsec:transversal-goal-hex}

Finally, we provide an alternative encoding with explicit boards and implicit goals.
This time, we specify the winning condition in the transversal game. This can
be applied to any positional game, but it is effective for Hex because specifying
that White has no connecting path is easier than specifying that Black has a 
connecting path. 

Below, $\borders$ and $\bordere$ denote White's starting and ending border.
We introduce variables $\breach{v}$, which hold for all positions
that are connected to $\borders$ through white and empty ($\neg\bblack{w}{d}$) positions only.
Then we simply check that no  $\bordere$ position is connected.
The following equations for ET replace the goal-detection clauses in EA or EN.
\begin{align}
& \exists \{\breach{v} \mid v \in \allnodes \} &
\end{align}

We get the following clauses
\begin{align}
& \neg \bblack{v}{d} \rightarrow \breach{v} & \text{for } v \in \borders\\
& \breach{v} \wedge \neg\bblack{w}{d} \rightarrow \breach{w} & \text{for } (v,w) \in \EDGES\\
& \neg \breach{v} & \text{for } v \in \bordere
\end{align}


\section{Implementation and Evaluation}
\label{sec:implementation}

We provide two new tools to generate explicit goal encodings for all maker-breaker games
and implicit goal encodings for Hex and HTTT.
The first tool is an extension of the COR encoding by \citet{10.1007/978-3-030-51825-7_31}, 
which allows an implicit goal specification and generates the EA, EN and ET encodings in DIMACS format.
The second tool\footnote{https://github.com/irfansha/Q-sage} generates the implicit board encodings LN, SN, LA and SA in QCIR format.
These can be translated to the QDIMACS format using the \citet{Tseitin1983CDPC} transformation.

\subsection{Size of the QBF encodings}
\label{subsec:qbfsize}

\begin{table}[t]
\centering
\caption{Alternation depth and size of explicit board encodings
(QDIMACS) and implicit board encodings (QCIR).}
\label{tab:encoding-size-dimacs-min}
\begin{tabular}{@{}lccl@{}}
\toprule
Enc. & Alt. & \# Variables & \multicolumn{1}{c}{\# Clauses}\\
\midrule
EA &  & $d\nbnodes+\size{\WIN}$ & $\frac{1}{2}d\nbnodes\lg\nbnodes+d \size{\WIN}$\\
EN & $d$ & $\frac{3}{2}d\nbnodes$ & $\frac{1}{2}d\nbnodes\lg\nbnodes$\\
ET &  & $d\nbnodes$ & $\frac{1}{2}d\nbnodes\lg\nbnodes+\size{\EDGES}$\\
%
\midrule
Enc. & Alt. & \# Variables & \multicolumn{1}{c}{\# Gates}\\
\midrule
LA            & \multirow{2}{*}{$d+1$} & \multirow{2}{*}{$\frac{3}{2}d\lg\nbnodes$} & $\frac{1}{2}d\nbnodes+\size{\WIN}$\\
LN      &  &  & $4d\lg\nbnodes+4\nbnodes$\\
\midrule
SA           & \multirow{2}{*}{$d$} & \multirow{2}{*}{$\frac{3}{2}d\lg\nbnodes$} & $d^2\ceil{\lg\nbnodes} + \frac{1}{2}d\nbnodes + \size{\WIN} $\\
SN      &  &  & \multicolumn{1}{l}{$d^2\ceil{\lg\nbnodes} + \frac{1}{2}d\nbnodes$} \\
\bottomrule
\end{tabular}
\end{table}

\begin{table*}
\centering
\caption{Size of symbolic goal encodings on an empty 19$\times$19 Hex board in QDIMACS (preprocessing time limit 30 min)}
\label{tab:encoding-size-hex-19}
\begin{tabular}{@{}lccccc@{\qquad}cccc@{}}
\toprule
Enc. & \multicolumn{4}{c}{Generated QBF Encodings (vars/cls)} && \multicolumn{4}{c}{After preprocessing with Bloqqer (vars/cls)}\\
\cmidrule(r){2-5} \cmidrule{7-10}
 & d=45 & d=91 & d=181 & d=361 && d=45 & d=91 & d=181 & d=361 \\
\midrule
EN  & 25k/122k & 50k/246k  & 100k/488k  & 199k/972k   && 25k/105k & 50k/242k  & 100k/510k & 199k/1044k \\
ET  & 17k/100k & 34k/200k  & 67k/395k   & 134k/785k   && 17k/100k & 34k/199k  & 67k/394k  & 134k/784k\\
LN  & 5k/22k   & 9k/33k    & 17k/54k    & 32k/97k     && 5k/7k    & 9k/12k    & 17k/21k   & 32k/39k\\
SN  & 53k/261k & 167k/720k & 560k/2182k & 2027k/7342k && 53k/122k & 167k/316k & TO        & TO \\
\bottomrule
\end{tabular}
\end{table*}

\pgfplotsset{
legend style={matrix anchor=north west, at={(0,0)}, font={\scriptsize}},
height=52mm,
width=1\linewidth,
}

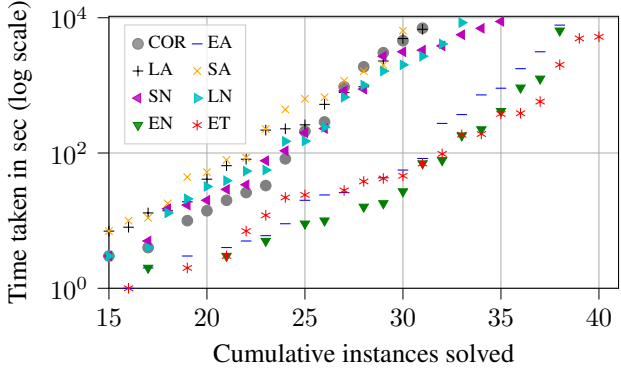
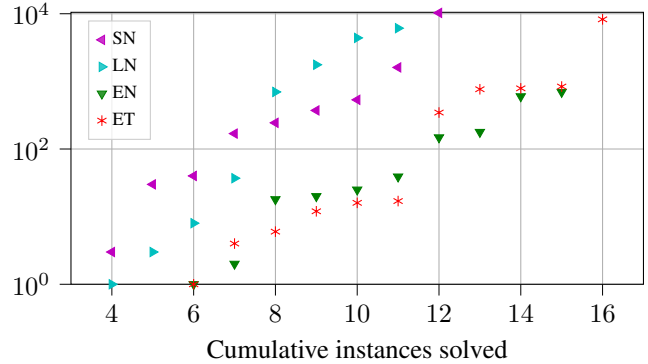
\begin{figure*}[t]
\centering
\begin{subfigure}[b]{0.47\textwidth}
\input{plots/all_hein_time}
\caption{Hein instances solved by our $7$ encodings and the original COR encoding.
$3$ hour time limit and $8$GB memory limit (log scale).}
\label{fig:allcactusplot}
\end{subfigure}
\hfill
\pgfplotsset{
width=1.1\linewidth,
}
\begin{subfigure}[b]{0.47\textwidth}
\centering
\hspace{-5mm}
\input{plots/all_champ_time}
\caption{Championship instances solved by $4$ implicit goal encodings.\\
$3$ hour time limit and $32$GB memory limit (log scale)}
\label{fig:allchampcactusplot}
\end{subfigure}
\caption{Experimental Results}
\end{figure*}

Table~\ref{tab:encoding-size-dimacs-min} shows
the alternation depth, number of variables, and number of clauses/gates for the explicit
and implicit board encodings of Hex,\footnote{We display here an asymptotically equivalent function. Appendix~\ref{sec:AnalysisEncsize} shows the exact size.}
depending on the number of positions ($\nbnodes$),
the depth of the game ($d$), the size of the winning set $\WIN$, and the number of edges $\EDGES$.
Obviously, the explicit-goal versions (*A) depend on $\size{\WIN}$, which is unpractical for 19$\times$19 Hex boards.
The implicit board versions (L*) are more concise than the explicit board versions (E*), 
at the expense of an extra quantifier alternation.
The stateless encoding saves this quantifier, but its size grows quadratically in the depth of the game, 
which is unpractical for long games.

In Table~\ref{tab:encoding-size-hex-19} we measured the actual number of variables and clauses
generated by our tools for increasingly deeper games on an empty 19$\times$19 board. 
These can only be generated for the implicit goal encodings (*N,*T).
Here we translated the QCIR benchmarks to QDIMACS for a fair comparison.
We show the counts before and after preprocessing the QBF with Bloqqer
(cf.~Sec.~\ref{subsec:experiments}).


Unfortunately, existing QBF solvers cannot solve any of these games on a 19$\times$19 board.
To study the effect of the encodings on the performance of QBF solvers, we experiment with
the Piet Hein benchmark of Hex puzzles on small boards. We also experiment with ``shallow'' end games
on a 19$\times$19 board, obtained from human championships. In the latter, after removing useless locations, the number
of open positions is quite small, giving a fair chance to all encodings.

\subsection{Solver Performance for Various Encodings}
\subsubsection{Experiment Design.}
\label{subsec:experiments}

We perform two experiments to evaluate the performance of QBF solvers on our encodings.
The first one is on a benchmark of 40 instances of Piet Hein's Hex puzzles from \citet{10.1007/978-3-030-51825-7_31}.
The second one runs on games from the best human players in the recent 19$\times$19 Hex championship.%
\footnote{Available at \url{https://www.littlegolem.net}. Appendix~\ref{sec:champbench} contains a detailed description of the benchmark selection.}
These games are resigned at quite an early point. This yields 15 UNSAT instances (for depths 11-17).
Then we finish the game using the Hex-specific solver Wolve \cite{arneson2009mohex} and roll back some moves.
This provides 8 SAT instances (for depths 9-17).
After pruning useless positions (cf.~Sec.~\ref{subsec:gex}),
the number of open positions in the first set of benchmarks is at most 40.
The 19$\times$19 boards have 10-100 open positions. Due to the filled positions and the
moderate game depth, many open positions become useless.

We first run many combinations of QBF preprocessors and solvers on all encodings for intermediate-size instances.
Here we only report the results on the strongest combination,\footnote{See Appendix \ref{sec:combprepsolvers} for discussion on other combinations.}
which is preprocessor Bloqqer \cite{DBLP:conf/cade/BiereLS11} with solver Caqe \cite{RabeT2015}.
We run Bloqqer/Caqe on all encodings for all benchmarks on a Huawei
FusionServer Pro V1288H V5 server with 384 GB memory and 3.0 GHz Intel Xeon Gold
6248R processor (using one core).
We measured the time and memory used.

\subsubsection{Experimental Results.}
\label{subsec:results}




Figure~\ref{fig:allcactusplot} shows a cactus plot of all Hein instances solved within the time and memory limits for all seven encodings.
Figure~\ref{fig:allchampcactusplot} shows the results for the championship benchmarks on the four encodings with implicit goal descriptions.%
\footnote{In Appendix~\ref{sec:heinmemoryplots}, we also present memory plots, and plots with SAT and UNSAT instances separated.
These plots don't modify the main messages.}
We first present our main findings:

(i) Clearly, in both experiments, explicit board encodings (E*) perform an order of magnitude faster than implicit board encodings (L*, S*).
So the extra quantifier alternation in L*-encodings is not compensated by their smaller size.

(ii) For Hein puzzles, we observe that each implicit neighbor encoding performs better than the corresponding explicit neighbor encoding.
(LN vs LA, SN vs SA, EN vs EA). For ET vs EA, this still holds for the harder instances. Note that for the second experiment,
on 19$\times$19 boards, all winning paths (*A) could not be enumerated practically.

(iii) Within the implicit board encodings, SN and LN perform quite similarly, and none of them dominates the other.
In the Hein instances, SN solves 2 UNSAT instances uniquely.
In the 19$\times$19 boards, SN solves 2 SAT instances uniquely and LN solves one UNSAT instance uniquely.
Overall, SN manages to solve a few more instances than LN.

(iv) ET is the only encoding that solves all 40 instances of the first experiment (up to depth 15 for UNSAT cases).
It solves most instances of the second experiment: 11 UNSAT (of which one uniquely) and 5 SAT instances.
Within the explicit board encodings, ET performs somewhat better than EN, but it doesn't clearly dominate it.
Apparently, checking that white can never win is easier than checking that black has won, but this may be
quite specific to Hex.

(v) Compared to the original COR encoding (rerun with Bloqqer/Caqe), the implicit board encodings LA and SA perform similarly,
whereas the implicit goal encodings (*N) perform better. The explicit board encodings (E*) on the other hand 
significantly outperform the COR encoding.

\subsubsection{Discussion.}
Although implicit board encodings didn't perform well in these experiments, we believe that the idea should not be dismissed.
Their strength is a more concise representation, but the difference only shows for larger board sizes and game depths
than we can currently solve. In a sense, pruning useless positions may have gained explicit board encodings (E*)
more advantage than implicit board encodings (L*). Similarly, the encodings that list all paths (*A) profit from the fact
that we listed only the minimal paths since otherwise, some instances would have had over 0.5M winning paths.
The neighbor-based versions (*N) have a disadvantage since they represent all winning paths.

Future QBF solvers might be able to better exploit the structure present in the more concise 
symbolic encodings. Also, future tools might improve the naive Tseitin circuit translation to clauses, on which the symbolic encodings (L*, S*) rely, while the explicit
encodings (E*) are generated in carefully handcrafted, optimal clauses.
The symbolic encoding could actually be made even more compact by compressing the witness path
logarithmically, at the expense of more quantifier alternations, following the iterative squaring technique in bounded model checking
and compact planning \citet{JUSSILA200745, DBLP:conf/ecai/CashmoreFG12}. 
This would only pay off for even deeper games.

Finally, we want to stress that, although the presented encoding ideas apply to all positional games, the trade-off between 
the various encodings may be quite different, depending considerably on the board size, 
the complexity of the moves and the winning conditions.

%

\section{Implicit Goals for Harary's Tic-Tac-Toe}
\label{sec:implicitgttt}
Harary's Generalized Tic-Tac-Toe (HTTT) is a positional game on an $n\times n$ board of squares, won by forming (a symmetric variant of)
a connected shape from a fixed, given set, called polyominos.
\citet{DBLP:conf/sat/DiptaramaYS16} provided a QBF encoding for solving the standard maker-maker version of HTTT.
Later, \citet{10.1007/978-3-030-51825-7_31} solved the proposed benchmarks efficiently with the COR encoding,
considering instances up to 4$\times$4 boards with polyominos of 4 cells.
\citet{BoucherV2021} specialized COR for HTTT, based on pairing, and consider 5$\times$5 boards.
The polyominos (which have funny standard names) can be characterized by two neighbor relations, up and right.
For example, \emph{Tippy} is a 4 cell polyomino, consisting of positions p0, p1, p2, p3, connected as right(p0, p1), up(p2, p1), right(p2, p3).

We adapt our SN encoding from Hex to the maker-breaker variant of HTTT easily, by using the two neighbor relations
\emph{right} and \emph{up}.
Each pair of neighbors in the witness sequence from the SN encoding is connected using the appropriate neighbor relation, depending on the polyomino.
In our encoding, symmetric variants of polyominos are simply treated as different ones.
We apply symmetry reduction, by restricting the position of the first move.
The extension is not specific to SN, but can easily be applied to the other implicit goal encodings as well.

We now generate SN encodings on a 5$\times$5 board.
We consider all 8 polyominos up to 4 cells and 3 other shapes: \emph{Z}, \emph{L} (both with 5 cells) and \emph{Snaky} (6 cells).
For each polyomino, we generate encodings of depth 7 to 15 (steps of 2). 
This gives 55 instances, encoded in a cumulative size of 2.8 MB.

We solved these instances with the same settings of Bloqqer/Caqe as before,
using a 3-hour time limit and 8 GB memory limit.
All polyominos up to 4 cells are solved within 10 seconds, except for \emph{Fatty} (which does not have a winning strategy and times out at depth 15).
The polyomino \emph{Z} is solved at depth 11 within 8 minutes and the other polyominos time out at depth 15. 
For all solved polyominos, the existence of a winning strategy of smaller depth is refuted within the time limit.

\jaco{Could compare better to \citet{BoucherV2021}, who treat Snaky on 5$\times$5 as well.}

\section{Conclusion and Future Work}
\label{sec:conclusionandfuturework}

We addressed two bottlenecks for the QBF encoding of maker-breaker positional games.
Using \emph{symbolic goal constraints} proved to be essential for generating QBF formulas
for games with many winning configurations, such as a standard Hex 19$\times$19 board.
On smaller boards, this technique also led to a boost in performance, leading to
currently the fastest way to solve Hex puzzles using a generic solver.

The other technique, using \emph{symbolic board positions}, leads to even smaller, lifted
QBF encodings, but current solvers cannot solve them as fast as the explicit board
representation. Still, many game instances could be solved with implicit board representation.
Some of our encodings have been submitted to the QBFeval competition, in the hope
that future QBF solvers might be able to exploit the special structure of our encodings.

On a Hex 19$\times$19 board, the game depth that can be handled is quite limited. 
This is not surprising, since game depth corresponds to QBF alternation depth, which
is the major factor of complexity for QBF solvers.
For limited game depth, many Hex board positions can be pruned away, which brings these
instances close to small puzzles. We also investigated game positions where human
champions resigned, because they believe the game is lost. Still, the actual number of
moves required from these lost positions is quite high. Experts use several patterns specific
to Hex (like ladders and bridges) to look ahead many steps. Even specialized Hex
solvers have difficulty solving these ``lost'' positions, but the specialized solvers
are of course much more efficient on Hex than generic QBF solvers on our encodings.

However, our approach is more general and is already applicable to any maker-breaker
positional game, as we demonstrated for Harary's Tic-Tac-Toe. 
The ideas can easily be applied to maker-maker variants of positional
games (for Hex, the two variants are equivalent). It is interesting for future work to apply
similar techniques to non-positional games, like Connect-Four and Breakthrough. 

\jaco{Missing references?}

\appendix

\section{Selection of Championship Benchmarks}
\label{sec:champbench}

Extending the experiment design in Section 5.2, We detail the selection of the Championship Human-played HEX games.

We considered all games among the top human players in the recent 19x19 Hex championship,
available at \href{https://www.littlegolem.net/jsp/tournament/tournament.jsp?trnid=hex.cv.HEX19.27.1.1}{littlegolem}.

These games are resigned at a quite early point. It can still be a challenge to finish those games to the end.
To get an impression, the reader can try them out through a GUI, for instance these games
played by top-players:
\begin{itemize}
\item \href{https://www.trmph.com/hex/game/lg-2291874}{game-2291874}
\item \href{https://www.trmph.com/hex/game/lg-2291880}{game-2291880}
\end{itemize}

We harvested games between the top players from the last championship.
Since the QBF solvers can only solve these games for limited depth, we
generated the encodings for depth 11-17; none of the games terminate within 17 steps,
so all these instances are UNSAT. Then we pruned useless positions  (see Section~\ref{subsec:gex}).
After pruning, in several cases no open positions were left.
We removed those games from the benchmark. As a result, we obtained 15 UNSAT instances of depth 11-17.
These are part of our benchmark.

To obtain some SAT instances as well, we need to finish the game. Here we used the Hex specific solver
Wolve \cite{arneson2009mohex}. Wolve could only solve a few of these games.
From the final won position, we rolled back 9-17 moves. After rolling back $n$ steps,
we generated instances for $n$ and $n+2$ steps. In the end, this procedure provided us
with 8 SAT instances of depth 9-17.



\section{Combinations QBF preprocessors / solvers}
\label{sec:combprepsolvers}
We now detail the experiments on the performance of all combinations of QBF preprocessors and solvers for our encodings
on a selection of the benchmark instances.

We tried all combinations of preprocessors and QBF solvers, both for QDIMACS and QCIR format
on all encodings for a selection of benchmarks (excluding very small and very large ones).
We now provide an overview of the tools we have experimented with.

\noindent
We have considered the following preprocessors:
\begin{itemize}
\item Bloqqer \cite{DBLP:conf/cade/BiereLS11}
\item HQSpre \cite{DBLP:journals/jsat/WimmerSB19}
\item QRATpre+ \cite{DBLP:conf/sat/LonsingE19}
\end{itemize}

We considered these QBF-solvers that operate directly on QCIR circuit format
(for those benchmarks that are generated in QCIR):
\begin{itemize}
\item CQuesto \cite{DBLP:conf/sat/Janota18}
\item Qfun \cite{DBLP:journals/corr/abs-1710-02198}
\item Qute, Quabs \cite{DBLP:journals/corr/Tentrup16}
\end{itemize}

We considered these QBF-solvers for QDIMACS format. We ran them both on encodings
generated in QDIMACS, and on encodings generated in QCIR, after a standard Tseitin
transformation to clauses.
\begin{itemize}
\item Caqe \cite{DBLP:conf/fmcad/RabeT15}
\item DepQBF \cite{DBLP:conf/cade/LonsingE17}
\item Qute \cite{DBLP:journals/jair/PeitlSS19}
\item Questo \cite{DBLP:conf/ijcai/JanotaM15}
\end{itemize}

Surprisingly, the QDIMACS solvers applied after a naive Tseitin transformation seem
to perform better than specialized QCIR solvers on the same encodings.

Overall, the combination of Bloqqer/Caqe performed the best,
although for some encodings, another preprocessor sometimes won.
We only report experiments with the best combination, Bloqqer/Caqe in Section \ref{subsec:results}


\section{Detailed Analysis of Encoding Sizes}
\label{sec:AnalysisEncsize}
Section \ref{subsec:qbfsize} contains an overview of the alternation depth and size of generated encodings,
in terms of the number of positions, game depth, and number of edges and winning configurations.
We provided simplified, asymptotically equivalent functions. This means that we only
gave the dominant terms, with the proper constants; we suppressed the smaller terms. For instance,
instead of $3n^2+2n$ we would have reported $3n^2$.

Here, we provide the precise functions. We do this both
for the EA, EN, ET encodings in DIMACS format (Table~\ref{tab:encoding-size-dimacs}) and
for the LA, LN, SA, SN encodings in QCIR (Table~\ref{tab:encoding-size-qcir}).
\jaco{Figure out the precise value(s) of constant $c$}
\jaco{Does separation and- and or-gates make sense? (given negations)}

In Table~\ref{tab:encoding-size-qcir}, we also show two new encodings, LI and SI, which
correspond to encodings with iterative squaring, as used in bounded model checking and classical planning
\citet{JUSSILA200745, DBLP:conf/ecai/CashmoreFG12}.
We have implemented these encodings, and we
mention these techniques in (see Section~\ref{subsec:results}). We suppressed
a detailed discussion because this technique only makes sense for deeper games (more moves)
than we can handle. However, the LI and SI encoding are interesting from a theoretical
perspective, because they encode the winning paths in logarithmic length, rather than linear length.
This gain comes at the cost of an increase in alternation depth.

\begin{table*}[p]
\centering
\caption{Size of the proposed explicit board (QDIMACS) encodings.}
\label{tab:encoding-size-dimacs}
\begin{tabular}{@{}llrrrrr@{}}
\toprule
\multicolumn{2}{c}{Encoding} & Altern. & Variables & \multicolumn{3}{c}{Clauses}\\
\cmidrule{1-2}\cmidrule{5-7}
Board & Goal                 & depth   &          & Binary & Ternary & Long \\
\midrule
Explicit &                   & $d$     & \multicolumn{1}{l}{$(d+1)\nbnodes+d\ceil{\lg\nbnodes})$} & \multicolumn{1}{l}{$d\nbnodes+\nbnodes\ceil{\lg\nbnodes}$} & \multicolumn{1}{l}{$\frac{d-1}{2}\nbnodes\ceil{\lg\nbnodes}$} & \multicolumn{1}{l}{$\frac{d-1}{2}\nbnodes$}\\
EA&       All winning        & & $+|\WIN|$ & $+\sum_{h \in \WIN} |h|$ & $+0$ & $+1$\\
EN&       Neighbors     & & $+\frac{d+1}{2}\nbnodes$ & $+\frac{d+1}{2}\nbnodes$ & $+0$ & $+1+(\nbnodes-|\bordere|)\ell$\\
ET&       Transversal          & & $+\nbnodes$ & $+|\borders|$ & $+|E|$ & $+0$\\
\bottomrule
\end{tabular}
\end{table*}


\begin{table*}[p]
\centering
\caption{Size of the proposed implicit board (QCIR) encodings.}
\label{tab:encoding-size-qcir}
\begin{tabular}{@{}llrrr@{}}
\toprule
\multicolumn{2}{c}{Encoding} & Altern. & Variables & Gates\\
\cmidrule{1-2}
Board & Goal                 & depth   &          & \\
\midrule
Lifted &                   & $d+1$     & \multicolumn{1}{l}{$(d+1)\ceil{\lg\nbnodes}+2d$} & \multicolumn{1}{l}{$2d(\ceil{\lg\nbnodes} + c)$}\\
LA&    All winning            & & $+\pathlength\ceil{\lg\nbnodes}$ & $+\pathlength(\nbnodes + 2\ceil{\lg\nbnodes} + 1) +2\nbnodes + 1 + |\WIN| $\\
LN&       Neighbors      & & $+(\pathlength+1)\ceil{\lg\nbnodes}$ & $+(4\pathlength-2)\ceil{\lg\nbnodes}+4\nbnodes + 2\pathlength + 2$\\
LI&       Iterative-sq.          & $+\ceil{\lg\pathlength}$ & $+(3\ceil{\lg\pathlength} + 2)\ceil{\lg\nbnodes} + \ceil{\lg\pathlength}$ & - \\
\midrule
Stateless &                   & $d$     & \multicolumn{1}{l}{$d\ceil{\lg\nbnodes}$} & $\ceil{\lg\nbnodes}((d-1)d + (d+1)\pathlength) + \frac{d^2+4d+3}{4}$ \\
SA&       All winning            & & $+\pathlength\ceil{\lg\nbnodes}$ & $+\pathlength\nbnodes +1 + |\WIN| $\\
SN&       Neighbors      & & $+(\pathlength)\ceil{\lg\nbnodes}$ & $+c\ceil{\lg\nbnodes}(\pathlength-1)+\nbnodes\pathlength$ \\
SI&       Iterative-sq.          & $+\ceil{\lg\pathlength}$ & $+(3\ceil{\lg\pathlength} + 2)\ceil{\lg\nbnodes} + \ceil{\lg\pathlength}$ & - \\
\bottomrule
\end{tabular}
\end{table*}

\section{Memory Plots for Hein's Hex Benchmarks}
\label{sec:heinmemoryplots}

In the Section \ref{subsec:results}, we presented cactus plots showing how many instances can be solved within a certain \emph{time} limit.
Figure~\ref{fig:allcactusplotmem} below shows  a similar cactus plot for the number of solved instances by Bloqqer/Caqe within a certain \emph{memory} limit, depending on our encodings.

\pgfplotsset{
legend pos={north west},
legend style={font={\scriptsize}},
height=50mm,
width=\linewidth,
}

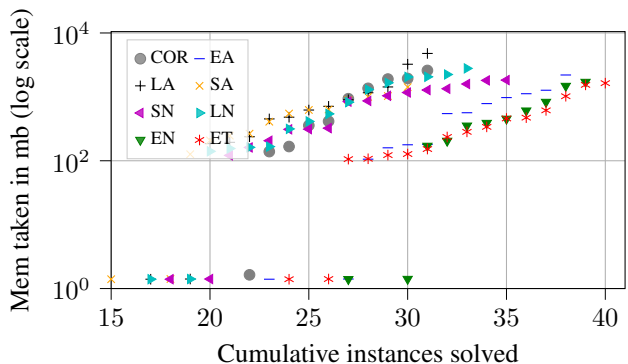
\begin{figure}[h]
\centering
\input{plots/all_hein_mem}
\caption{Cumulative Hein instances solved by $7$ encodings within $3$ hours.
Dependence on memory limit (log scale)}
\label{fig:allcactusplotmem}
\end{figure}

Similar to the time-based analysis, the QBF solver combination Bloqqer/Caqe uses the
least memory for the explicit board encodings with implicit goals ET and EN, followed by EA (explicit goals).
Next are the symbolic encodings with implicit goal conditions LN and SN. Finally, worst are the symbolic
board encodings with explicit goal conditions (LA and SA).

\section{Separation of SAT and UNSAT instances}
\label{sec:satandunsat}

\subsection{Hein's Benchmarks, separated}

Here we separate the cactus plot for all Hein's Hex benchmarks in SAT instances and UNSAT instances.
We investigate if different encodings would handle SAT or UNSAT instances better.
This would be interesting for portfolio approaches.

Figure~\ref{fig:heinsatunsat} shows the separation in UNSAT (left) and SAT (right) instances for Hein's benchmarks.
We show the dependence on time (top) and memory (bottom).
The pictures show basically the same trends, so the separation
between SAT and UNSAT cases reveals no new information.

\jaco{Do proper statistical tests on SAT versus UNSAT}

\begin{figure*}[p]
\begin{subfigure}[b]{0.47\textwidth}
\centering
\input{plots/uns_hein_time}
\caption{UNSAT instances (time)}
\label{fig:unsatcactusplottime}
\end{subfigure}
\hfill
\begin{subfigure}[b]{0.47\textwidth}
\centering
\input{plots/sat_hein_time}
\caption{SAT instances (time)}
\label{fig:satcactusplottime}
\end{subfigure}
\begin{subfigure}[b]{0.47\textwidth}
\centering
\input{plots/uns_hein_mem}
\caption{UNSAT instances (memory)}
\label{fig:unsatcactusplotmem}
\end{subfigure}
\hfill
\begin{subfigure}[b]{0.47\textwidth}
\centering
\input{plots/sat_hein_mem}
\caption{SAT instances (memory)}
\label{fig:satcactusplotmem}
\end{subfigure}
\caption{Cumulative Hein instances solved by $7$ encodings with $3$ hour time limit and $8$GB memory limit (log scale)}
\label{fig:heinsatunsat}
\end{figure*}

\subsection{Championship Benchmarks, separated}
Here we separate all Championship Hex benchmarks in cactus plots for the SAT
instances and for the UNSAT instances.

The results are shown in Figure~\ref{fig:championshipsatunsat}. We display UNSAT instances
on the left and SAT instances on the right. The top row focuses on time,
and the bottom row on memory.
Also in this case, the separation between SAT and UNSAT instances does not provide deviating insights.

\pgfplotsset{
legend pos={north west},
legend style={font={\scriptsize}},
height=54mm,
width=0.9\linewidth,
}

\begin{figure*}[t]
\begin{subfigure}[b]{0.47\textwidth}
\centering
\input{plots/uns_champ_time}
\caption{UNSAT instances (time)}
\label{fig:champunsatcactusplottime}
\end{subfigure}
\hfill
\begin{subfigure}[b]{0.47\textwidth}
\centering
\input{plots/sat_champ_time}
\caption{SAT instances (time)}
\label{fig:champsatcactusplottime}
\end{subfigure}
\begin{subfigure}[b]{0.47\textwidth}
\centering
\input{plots/uns_champ_mem}
\caption{UNSAT instances (memory)}
\label{fig:champunsatcactusplotmem}
\end{subfigure}
\hfill
\begin{subfigure}[b]{0.47\textwidth}
\centering
\input{plots/sat_champ_mem}
\caption{SAT instances (memory)}
\label{fig:champsatcactusplot}
\end{subfigure}
\caption{Cumulative championship instances solved by $4$ encodings with $3$ hour time limit and $32$GB memory limit (log scale)}
\label{fig:championshipsatunsat}
\end{figure*}

\bibliography{references}

\end{document}

%% file: commands.tex
\usepackage{aaai23}  
\usepackage{times}  
\usepackage{helvet}  
\usepackage{courier}  
\usepackage[hyphens]{url}  
\usepackage{graphicx} 
\usepackage{natbib}  
\usepackage{caption} 
\DeclareCaptionStyle{ruled}{labelfont=normalfont,labelsep=colon,strut=off} 
\nocopyright
\usepackage{hyperref}

\makeatletter
\def\orcidID#1{\smash{\href{https://orcid.org/#1}{\protect\raisebox{-1.25pt}{\protect\includegraphics{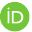}}}}}
\makeatother

\usepackage{tikz}
\usetikzlibrary{arrows,snakes}
\usetikzlibrary[calc,shapes,patterns,positioning,decorations.pathmorphing,decorations.pathreplacing]
\usepackage{pgfplots}
\DeclareUnicodeCharacter{2212}{−}
\usepgfplotslibrary{groupplots,dateplot}
\pgfplotsset{compat=newest}

\definecolor{light-gray}{gray}{0.92}
\tikzset{
  hav-circl/.style={draw,circle,minimum size=3.5mm,inner sep=0,text=black,font=\scriptsize},
  hav-white/.style={hav-circl,fill=light-gray},
  hav-black/.style={hav-circl,,fill=black,text=white},
  hav-noone/.style={text=black,font=\scriptsize},
  hav-empty/.style={shape=regular polygon,regular polygon sides=6,draw,minimum size=5.85mm,inner sep=0,outer sep=0,font=\scriptsize},
}
\newcommand{\havcoordinate}[4]{
  \foreach \i in {#1,...,#2} {
    \foreach \j in {#3,...,#4} {
      \coordinate (\i-\j) at ({(\i+1) * 1.732},{(\j)});
    }
  }
  \foreach \i in {#1,...,#2} {
    \foreach \j in {#3,...,#4} {
      \coordinate (\i--\j) at ({(\i+1+1/2) * 1.732},{(\j+1/2)});
    }
  }
}
\newcommand{\scala}{0.57}

\usepackage{subcaption}

%% file: hein9.tex
\subcaptionbox{Original puzzle\label{fig:9-original}}{ 
  \begin{tikzpicture}[>=stealth',scale=\scala,every node/.style={scale=2*\scala}]
  \havcoordinate{0}{5}{0}{5}
  \node[hav-white] at ([shift={(-0.3,-0.3)}]1--1) {}; \node[hav-black] at ([shift={(-0.3,0.3)}]1--4) {};
  \node[hav-black] at ([shift={(0.3,-0.3)}]3--1) {}; \node[hav-white] at ([shift={(0.3,0.3)}]3--4) {};
  \node[hav-empty] at  (1-3) {$a_1$};
  \node[hav-empty] at (1--2) {$a_2$};
  \node[hav-empty] at (1--3) {};
  \node[hav-empty] at  (2-2) {};
  \node[hav-empty] at  (2-3) {$b_2$};
  \node[hav-empty] at  (2-4) {};
  \node[hav-empty] at (2--1) {};
  \node[hav-empty] at (2--2) {$b_3$};
  \node[hav-empty] at (2--3) {$c_2$};
  \node[hav-empty] at (2--4) {$d_1$};
  \node[hav-empty] at  (3-2) {$b_4$};
  \node[hav-empty] at  (3-3) {$c_3$};
  \node[hav-empty] at  (3-4) {$d_2$};
  \node[hav-empty] at (3--2) {$c_4$};
  \node[hav-empty] at (3--3) {$d_3$};
  \node[hav-empty] at  (4-3) {};
  \node[hav-black] at  (2-2) {};
  \node[hav-black] at (1--3) {};
  \node[hav-white] at  (2-4) {};
  \node[hav-white] at (2--1) {};
  \node[hav-white] at  (4-3) {};
  \end{tikzpicture}
}\hfill
\subcaptionbox{Filled-in Short Hex instance for $d=5$\label{fig:9-short-hex}}{
  \begin{tikzpicture}[>=stealth',scale=\scala,every node/.style={scale=2*\scala}]
  \havcoordinate{0}{5}{0}{5}
  \node[hav-white] at ([shift={(-0.3,-0.3)}]1--1) {}; \node[hav-black] at ([shift={(-0.3,0.3)}]1--4) {};
  \node[hav-black] at ([shift={(0.3,-0.3)}]3--1) {}; \node[hav-white] at ([shift={(0.3,0.3)}]3--4) {};
  \node[hav-empty] at  (1-3) {};
  \node[hav-empty] at (1--2) {};
  \node[hav-empty] at (1--3) {};
  \node[hav-empty] at  (2-2) {};
  \node[hav-empty] at  (2-3) {};
  \node[hav-empty] at  (2-4) {};
  \node[hav-empty] at (2--1) {};
  \node[hav-empty] at (2--2) {};
  \node[hav-empty] at (2--3) {};
  \node[hav-empty] at (2--4) {};
  \node[hav-empty] at  (3-2) {};
  \node[hav-empty] at  (3-3) {};
  \node[hav-empty] at  (3-4) {};
  \node[hav-empty] at (3--2) {};
  \node[hav-empty] at (3--3) {};
  \node[hav-empty] at  (4-3) {};
  \node[hav-black] at  (2-2) {};
  \node[hav-black] at (1--3) {};
  \node[hav-white] at  (1-3) {};
  \node[hav-white] at  (2-4) {};
  \node[hav-white] at (2--1) {};
  \node[hav-white] at  (4-3) {};
  \node[hav-white] at (2--3) {};
  \node[hav-white] at (2--4) {};
  \node[hav-white] at  (3-3) {};
  \node[hav-white] at  (3-4) {};
  \node[hav-white] at (3--2) {};
  \node[hav-white] at (3--3) {};
  \end{tikzpicture}
}\\
\subcaptionbox{Contracted Generalized Hex instance\label{fig:9-gex}}{
  \begin{tikzpicture}[>=stealth',scale=\scala,every node/.style={scale=2*\scala}]
  \havcoordinate{0}{5}{0}{5}
  \node[hav-noone] (s) at ([shift={(-0.3,0.3)}]1--4) {$s$}; \node[hav-noone] (e) at ([shift={(0.3,-0.3)}]3--1) {$e$};
  \phantom{\node[hav-empty] at  (1-3) {};} \phantom{\node[hav-empty] at  (4-3) {};}
  \node[hav-circl] (a2) at (1--2) {$a_2$};
  \node[hav-circl] (b2) at  (2-3) {$b_2$};
  \node[hav-circl] (b3) at (2--2) {$b_3$};
  \node[hav-circl] (b4) at  (3-2) {$b_4$};
  \draw (s)  -- (a2);
  \draw (s)  -- (b2);
  \draw (a2) -- (b3);
  \draw (b2) -- (b3);
  \draw (b3) -- (b4);
  \draw (b4) -- (e);
  \end{tikzpicture}
}\hfill
\subcaptionbox{Contracted Generalized Hex transversal instance\label{fig:9-transversal-gex}}{
  \begin{tikzpicture}[>=stealth',scale=\scala,every node/.style={scale=2*\scala}]
  \havcoordinate{0}{5}{0}{5}
  \node[hav-noone] (e) at ([shift={(-0.3,-0.3)}]1--1) {$e$}; \node[hav-noone] (s) at ([shift={(0.3,0.3)}]3--4) {$s$};
  \phantom{\node[hav-empty] at  (1-3) {};} \phantom{\node[hav-empty] at  (4-3) {};}
  \node[hav-circl] (a2) at (1--2) {$a_2$};
  \node[hav-circl] (b2) at  (2-3) {$b_2$};
  \node[hav-circl] (b3) at (2--2) {$b_3$};
  \node[hav-circl] (b4) at  (3-2) {$b_4$};

  \draw (s)  -- (b2);
  \draw (s)  -- (b3);
  \draw (s)  -- (b4);
  \draw (a2) -- (b2);
  \draw (a2) --  (e);
  \draw (b3) --  (e);
  \draw (b4) --  (e);
  \end{tikzpicture}
}

%% file: plots/all_hein_time.tex
\begin{tikzpicture}

\definecolor{darkgray176}{RGB}{176,176,176}
\definecolor{darkturquoise0191191}{RGB}{0,191,191}
\definecolor{darkviolet1910191}{RGB}{191,0,191}
\definecolor{gray}{RGB}{128,128,128}
\definecolor{green01270}{RGB}{0,127,0}
\definecolor{lightgray204}{RGB}{204,204,204}
\definecolor{orange}{RGB}{255,165,0}

\begin{axis}[
legend cell align={left},
legend columns=2,
legend style={
  fill opacity=0.8,
  draw opacity=1,
  text opacity=1,
  at={(0.03,0.97)},
  anchor=north west,
  draw=lightgray204
},
log basis y={10},
tick align=outside,
tick pos=left,
x grid style={darkgray176},
xlabel={Cumulative instances solved},
xmajorgrids,
xmin=15, xmax=41,
xtick style={color=black},
y grid style={darkgray176},
ylabel={Time taken in sec (log scale)},
ymajorgrids,
ymin=1, ymax=10500,
ymode=log,
ytick style={color=black}
]
\addplot [draw=gray, fill=gray, mark=*, only marks]
table{%
x  y
12 0.1
13 1
14 2
15 3
17 4
19 10
20 14
21 20
22 26
23 33
24 82
25 210
26 288
27 953
28 1884
29 3038
30 4581
31 7003
};
\addlegendentry{COR}
\addplot [draw=blue, fill=blue, mark=-, only marks]
table{%
x  y
13 0.1
16 1
17 2
19 3
21 4
22 5
23 6
24 9
25 20
26 24
27 26
29 43
30 56
31 83
32 274
33 370
34 723
35 908
36 1766
37 3138
38 7782
};
\addlegendentry{EA}
\addplot [draw=black, fill=black, mark=+, only marks]
table{%
x  y
5 0.1
10 1
11 3
13 4
14 6
15 7
16 8
17 13
18 14
19 19
20 41
21 65
22 81
23 218
24 228
25 261
26 524
27 786
28 953
29 2290
30 4927
31 6748
};
\addlegendentry{LA}
\addplot [draw=orange, fill=orange, mark=x, only marks]
table{%
x  y
4 0.1
10 1
11 3
12 5
13 6
15 7
16 10
17 11
18 18
19 44
20 52
21 79
22 85
23 226
24 440
25 633
26 667
27 1171
28 1628
29 2000
30 6436
};
\addlegendentry{SA}
\addplot [
  draw=darkviolet1910191,
  fill=darkviolet1910191,
  mark options={rotate=90},
  mark=triangle*,
  only marks
]
table{%
x  y
11 0.1
13 2
15 3
17 5
18 15
19 17
20 20
21 29
22 34
23 77
24 108
25 199
26 230
27 827
28 881
29 2692
30 3138
31 3316
32 3838
33 5611
34 6973
35 8814
};
\addlegendentry{SN}
\addplot [
  draw=darkturquoise0191191,
  fill=darkturquoise0191191,
  mark options={rotate=270},
  mark=triangle*,
  only marks
]
table{%
x  y
10 0.1
12 1
13 2
15 3
17 4
18 13
19 21
20 32
21 39
22 54
23 56
24 149
25 150
26 241
27 669
28 1007
29 1628
30 2014
31 2698
32 4102
33 8463
};
\addlegendentry{LN}
\addplot [draw=green01270, fill=green01270, mark options={rotate=180}, mark=triangle*, only marks]
table{%
x  y
11 0.1
14 1
17 2
21 3
23 5
25 9
26 10
28 16
29 18
30 27
31 70
32 78
33 181
34 223
35 416
36 923
37 1253
38 6436
39 10681
};
\addlegendentry{EN}
\addplot [draw=red, fill=red, mark=asterisk, only marks]
table{%
x  y
13 0.1
16 1
19 2
21 3
22 7
23 12
24 22
25 24
27 28
28 38
29 42
30 46
31 70
32 97
33 186
34 191
35 378
36 387
37 576
38 2016
39 4947
40 5231
};
\addlegendentry{ET}
\end{axis}

\end{tikzpicture}

%% file: plots/all_champ_time.tex
\begin{tikzpicture}

\definecolor{darkgray176}{RGB}{176,176,176}
\definecolor{darkturquoise0191191}{RGB}{0,191,191}
\definecolor{darkviolet1910191}{RGB}{191,0,191}
\definecolor{green01270}{RGB}{0,127,0}
\definecolor{lightgray204}{RGB}{204,204,204}

\begin{axis}[
legend cell align={left},
legend style={
  fill opacity=0.8,
  draw opacity=1,
  text opacity=1,
  at={(0.03,0.97)},
  anchor=north west,
  draw=lightgray204
},
log basis y={10},
tick align=outside,
tick pos=left,
x grid style={darkgray176},
xlabel={Cumulative instances solved},
xmajorgrids,
xmin=3, xmax=17,
xtick style={color=black},
y grid style={darkgray176},
ymajorgrids,
ymin=1, ymax=10500,
ymode=log,
ytick style={color=black}
]
\addplot [
  draw=darkviolet1910191,
  fill=darkviolet1910191,
  mark options={rotate=90},
  mark=triangle*,
  only marks
]
table{%
x  y
2 0.1
4 3
5 30
6 40
7 169
8 244
9 372
10 534
11 1608
12 10290
};
\addlegendentry{SN}
\addplot [
  draw=darkturquoise0191191,
  fill=darkturquoise0191191,
  mark options={rotate=270},
  mark=triangle*,
  only marks
]
table{%
x  y
3 0.1
4 1
5 3
6 8
7 37
8 698
9 1761
10 4394
11 6114
};
\addlegendentry{LN}
\addplot [draw=green01270, fill=green01270, mark options={rotate=180}, mark=triangle*, only marks]
table{%
x  y
5 0.1
6 1
7 2
8 18
9 20
10 25
11 39
12 148
13 178
14 599
15 695
};
\addlegendentry{EN}
\addplot [draw=red, fill=red, mark=asterisk, only marks]
table{%
x  y
5 0.1
6 1
7 4
8 6
9 12
10 16
11 17
12 346
13 765
14 789
15 839
16 8264
};
\addlegendentry{ET}
\end{axis}

\end{tikzpicture}

%% file: plots/all_hein_mem.tex
\begin{tikzpicture}

\definecolor{darkgray176}{RGB}{176,176,176}
\definecolor{darkturquoise0191191}{RGB}{0,191,191}
\definecolor{darkviolet1910191}{RGB}{191,0,191}
\definecolor{green01270}{RGB}{0,127,0}
\definecolor{lightgray204}{RGB}{204,204,204}
\definecolor{orange}{RGB}{255,165,0}

\begin{axis}[
legend cell align={left},
legend columns=2,
legend style={
  fill opacity=0.8,
  draw opacity=1,
  text opacity=1,
  at={(0.03,0.97)},
  anchor=north west,
  draw=lightgray204
},
log basis y={10},
tick align=outside,
tick pos=left,
x grid style={darkgray176},
xlabel={Cumulative instances solved},
xmajorgrids,
xmin=15, xmax=41,
xtick style={color=black},
y grid style={darkgray176},
ylabel={Mem taken in mb (log scale)},
ymajorgrids,
ymin=1, ymax=10500,
ymode=log,
ytick style={color=black}
]
\addplot [draw=gray, fill=gray, mark=*, only marks]
table{%
x  y
1 1.38
10 1.39
12 1.63
22 1.64
23 139.15
24 167.03
25 357.39
26 415.04
27 935.95
28 1360.0
29 1900.0
30 1940.0
31 2590.0
};
\addlegendentry{COR}

\addplot [draw=blue, fill=blue, mark=-, only marks]
table{%
x  y
23 1.4
27 1.41
28 105.83
29 159.47
30 177.87
31 180.02
32 546.98
33 564.38
34 784.53
35 973.97
36 1120
37 1270
38 2200
};
\addlegendentry{EA}
\addplot [draw=black, fill=black, mark=+, only marks]
table{%
x  y
17 1.4
19 1.41
20 185.45
21 194.23
22 237.83
23 451.88
24 480.68
25 622.19
26 710.95
27 901.27
28 1160
29 1430
30 3240
31 4780
};
\addlegendentry{LA}
\addplot [draw=orange, fill=orange, mark=x, only marks]
table{%
x  y
15 1.4
18 1.41
19 126.52
20 208.38
21 247.32
22 265.42
23 408.73
24 548.23
25 623.82
26 634.12
27 889.03
28 1000
29 1030
30 1420
};
\addlegendentry{SA}
\addplot [
  draw=darkviolet1910191,
  fill=darkviolet1910191,
  mark options={rotate=90},
  mark=triangle*,
  only marks
]
table{%
x  y
18 1.4
20 1.41
21 121.34
22 161.57
23 206.02
24 310.18
25 312.37
26 323.86
27 842.8
28 865.09
29 1040
30 1170
31 1280
32 1340
33 1590
34 1810
35 1830
};
\addlegendentry{SN}
\addplot [
  draw=darkturquoise0191191,
  fill=darkturquoise0191191,
  mark options={rotate=270},
  mark=triangle*,
  only marks
]
table{%
x  y
17 1.4
19 1.41
20 140.54
21 155.56
22 162.25
23 166.44
24 315.19
25 412.62
26 541.75
27 825.58
28 1300
29 1660
30 2050
31 2060
32 2240
33 2790
};
\addlegendentry{LN}
\addplot [draw=green01270, fill=green01270, mark options={rotate=180}, mark=triangle*, only marks]
table{%
x  y
27 1.4
30 1.41
31 170.85
32 204.53
33 353.53
34 389.68
35 452.94
36 608.6
37 832
38 1480
39 1710
};
\addlegendentry{EN}
\addplot [draw=red, fill=red, mark=asterisk, only marks]
table{%
x  y
24 1.4
26 1.41
27 106.08
28 107.43
29 122.61
30 127.52
31 152.56
32 235.64
33 283.31
34 339.07
35 458.98
36 473.02
37 615.05
38 1023.28
39 1540
40 1640
};
\addlegendentry{ET}
\end{axis}

\end{tikzpicture}

%% file: plots/uns_hein_time.tex
\begin{tikzpicture}

\definecolor{darkgray176}{RGB}{176,176,176}
\definecolor{darkturquoise0191191}{RGB}{0,191,191}
\definecolor{darkviolet1910191}{RGB}{191,0,191}
\definecolor{green01270}{RGB}{0,127,0}
\definecolor{lightgray204}{RGB}{204,204,204}
\definecolor{orange}{RGB}{255,165,0}

\begin{axis}[
legend cell align={left},
legend columns=2,
legend style={
  fill opacity=0.8,
  draw opacity=1,
  text opacity=1,
  at={(0.03,0.97)},
  anchor=north west,
  draw=lightgray204
},
log basis y={10},
tick align=outside,
tick pos=left,
x grid style={darkgray176},
xlabel={Cumulative instances solved},
xmajorgrids,
xmin=5, xmax=24,
xtick style={color=black},
y grid style={darkgray176},
ylabel={Time taken in sec (log scale)},
ymajorgrids,
ymin=1, ymax=10000,
ymode=log,
ytick style={color=black}
]
\addplot [draw=gray, fill=gray, mark=*, only marks]
table{%
x  y
9 0.1
10 2.0
11 3.0
12 33.0
13 82.0
14 210.0
15 288.0
16 7003.0
};
\addlegendentry{COR}
\addplot [draw=blue, fill=blue, mark=-, only marks]
table{%
x  y
9 0.1
11 1
12 4
13 24
14 26
15 43
16 56
17 274
18 370
19 908
20 1766
21 3138
22 7782
};
\addlegendentry{EA}
\addplot [draw=black, fill=black, mark=+, only marks]
table{%
x  y
3 0.1
7 1
8 3
9 6
10 7
11 19
12 81
13 228
14 261
15 786
16 953
};
\addlegendentry{LA}
\addplot [draw=orange, fill=orange, mark=x, only marks]
table{%
x  y
3 0.1
7 1
8 3
10 7
11 44
12 226
13 440
14 633
15 1171
16 2000
};
\addlegendentry{SA}
\addplot [
  draw=darkviolet1910191,
  fill=darkviolet1910191,
  mark options={rotate=90},
  mark=triangle*,
  only marks
]
table{%
x  y
8 0.1
9 3
10 5
11 20
12 77
13 108
14 199
15 230
16 827
17 5611
18 6973
19 8814
};
\addlegendentry{SN}
\addplot [
  draw=darkturquoise0191191,
  fill=darkturquoise0191191,
  mark options={rotate=270},
  mark=triangle*,
  only marks
]
table{%
x  y
7 0.1
8 1
9 3
10 4
11 21
12 54
13 56
14 149
15 241
16 1007
17 4102
};
\addlegendentry{LN}
\addplot [draw=green01270, fill=green01270, mark options={rotate=180}, mark=triangle*, only marks]
table{%
x  y
8 0.1
9 1
11 2
12 3
13 5
14 9
15 16
16 27
17 70
18 223
19 416
20 923
21 1253
22 6436
23 10681
};
\addlegendentry{EN}
\addplot [draw=red, fill=red, mark=asterisk, only marks]
table{%
x  y
8 0.1
10 1
11 2
12 12
13 22
14 24
15 28
16 70
17 97
18 186
19 378
20 387
21 576
22 2016
23 5231
};
\addlegendentry{ET}
\end{axis}

\end{tikzpicture}

%% file: plots/sat_hein_time.tex
\begin{tikzpicture}

\definecolor{darkgray176}{RGB}{176,176,176}
\definecolor{darkturquoise0191191}{RGB}{0,191,191}
\definecolor{darkviolet1910191}{RGB}{191,0,191}
\definecolor{green01270}{RGB}{0,127,0}
\definecolor{lightgray204}{RGB}{204,204,204}
\definecolor{orange}{RGB}{255,165,0}

\begin{axis}[
legend cell align={left},
legend columns=2,
legend style={
  fill opacity=0.8,
  draw opacity=1,
  text opacity=1,
  at={(0.03,0.97)},
  anchor=north west,
  draw=lightgray204
},
log basis y={10},
tick align=outside,
tick pos=left,
x grid style={darkgray176},
xlabel={Cumulative instances solved},
xmajorgrids,
xmin=5, xmax=18,
xtick style={color=black},
y grid style={darkgray176},
ymajorgrids,
ymin=1, ymax=10000,
ymode=log,
ytick style={color=black}
]
\addplot [draw=gray, fill=gray, mark=*, only marks]
table{%
x  y
3 0.1
4 1.0
6 4.0
8 10.0
9 14.0
10 20.0
11 26.0
12 953.0
13 1884.0
14 3038.0
15 4581.0
};
\addlegendentry{COR}
\addplot [draw=blue, fill=blue, mark=-, only marks]
table{%
x  y
4 0.1
5 1
6 2
8 3
9 4
10 5
11 6
12 9
13 20
14 43
15 83
16 723
};
\addlegendentry{EA}
\addplot [draw=black, fill=black, mark=+, only marks]
table{%
x  y
2 0.1
3 1
5 4
6 8
7 13
8 14
9 41
10 65
11 218
12 524
13 2290
14 4927
15 6748
};
\addlegendentry{LA}
\addplot [draw=orange, fill=orange, mark=x, only marks]
table{%
x  y
1 0.1
3 1
4 5
5 6
6 10
7 11
8 18
9 52
10 79
11 85
12 667
13 1628
14 6436
};
\addlegendentry{SA}
\addplot [
  draw=darkviolet1910191,
  fill=darkviolet1910191,
  mark options={rotate=90},
  mark=triangle*,
  only marks
]
table{%
x  y
3 0.1
5 2
6 3
7 5
8 15
9 17
10 29
11 34
12 881
13 2692
14 3138
15 3316
16 3838
};
\addlegendentry{SN}
\addplot [
  draw=darkturquoise0191191,
  fill=darkturquoise0191191,
  mark options={rotate=270},
  mark=triangle*,
  only marks
]
table{%
x  y
3 0.1
4 1
5 2
6 3
7 4
8 13
9 32
10 39
11 150
12 669
13 1628
14 2014
15 2698
16 8463
};
\addlegendentry{LN}
\addplot [draw=green01270, fill=green01270, mark options={rotate=180}, mark=triangle*, only marks]
table{%
x  y
3 0.1
5 1
6 2
9 3
10 5
11 9
12 10
13 16
14 18
15 78
16 181
};
\addlegendentry{EN}
\addplot [draw=red, fill=red, mark=asterisk, only marks]
table{%
x  y
5 0.1
6 1
8 2
10 3
11 7
12 28
13 38
14 42
15 46
16 191
17 4947
};
\addlegendentry{ET}
\end{axis}

\end{tikzpicture}

%% file: plots/uns_hein_mem.tex
\begin{tikzpicture}

\definecolor{darkgray176}{RGB}{176,176,176}
\definecolor{darkturquoise0191191}{RGB}{0,191,191}
\definecolor{darkviolet1910191}{RGB}{191,0,191}
\definecolor{green01270}{RGB}{0,127,0}
\definecolor{lightgray204}{RGB}{204,204,204}
\definecolor{orange}{RGB}{255,165,0}

\begin{axis}[
legend cell align={left},
legend columns=2,
legend style={
  fill opacity=0.8,
  draw opacity=1,
  text opacity=1,
  at={(0.03,0.97)},
  anchor=north west,
  draw=lightgray204
},
log basis y={10},
tick align=outside,
tick pos=left,
x grid style={darkgray176},
xlabel={Cumulative instances solved},
xmajorgrids,
xmin=5, xmax=24,
xtick style={color=black},
y grid style={darkgray176},
ylabel={Mem taken in mb (log scale)},
ymajorgrids,
ymin=1, ymax=10000,
ymode=log,
ytick style={color=black}
]
\addplot [draw=gray, fill=gray, mark=*, only marks]
table{%
x  y
1 1.38
3 1.39
4 1.63
11 1.64
12 139.15
13 167.03
14 357.39
15 415.04
16 1940.0
};
\addlegendentry{COR}
\addplot [draw=blue, fill=blue, mark=-, only marks]
table{%
x  y
12 1.4
14 1.41
15 159.47
16 180.02
17 546.98
18 564.38
19 784.53
20 1120
21 1270
22 2200
};
\addlegendentry{EA}
\addplot [draw=black, fill=black, mark=+, only marks]
table{%
x  y
9 1.4
11 1.41
12 237.83
13 451.88
14 480.68
15 901.27
16 1160
};
\addlegendentry{LA}
\addplot [draw=orange, fill=orange, mark=x, only marks]
table{%
x  y
8 1.4
10 1.41
11 126.52
12 408.73
13 548.23
14 623.82
15 1000
16 1030
};
\addlegendentry{SA}
\addplot [
  draw=darkviolet1910191,
  fill=darkviolet1910191,
  mark options={rotate=90},
  mark=triangle*,
  only marks
]
table{%
x  y
10 1.4
11 1.41
12 206.02
13 310.18
14 312.37
15 323.86
16 865.09
17 1340
18 1810
19 1830
};
\addlegendentry{SN}
\addplot [
  draw=darkturquoise0191191,
  fill=darkturquoise0191191,
  mark options={rotate=270},
  mark=triangle*,
  only marks
]
table{%
x  y
9 1.4
11 1.41
12 155.56
13 166.44
14 315.19
15 541.75
16 1660
17 2060
};
\addlegendentry{LN}
\addplot [draw=green01270, fill=green01270, mark options={rotate=180}, mark=triangle*, only marks]
table{%
x  y
15 1.4
16 1.41
17 204.53
18 389.68
19 452.94
20 608.6
21 832
22 1480
23 1710
};
\addlegendentry{EN}
\addplot [draw=red, fill=red, mark=asterisk, only marks]
table{%
x  y
14 1.4
15 1.41
16 152.56
17 235.64
18 283.31
19 458.98
20 473.02
21 615.05
22 1023.28
23 1540
};
\addlegendentry{ET}
\end{axis}

\end{tikzpicture}

%% file: plots/sat_hein_mem.tex
\begin{tikzpicture}

\definecolor{darkgray176}{RGB}{176,176,176}
\definecolor{darkturquoise0191191}{RGB}{0,191,191}
\definecolor{darkviolet1910191}{RGB}{191,0,191}
\definecolor{green01270}{RGB}{0,127,0}
\definecolor{lightgray204}{RGB}{204,204,204}
\definecolor{orange}{RGB}{255,165,0}

\begin{axis}[
legend cell align={left},
legend columns=2,
legend style={
  fill opacity=0.8,
  draw opacity=1,
  text opacity=1,
  at={(0.03,0.97)},
  anchor=north west,
  draw=lightgray204
},
log basis y={10},
tick align=outside,
tick pos=left,
x grid style={darkgray176},
xlabel={Cumulative instances solved},
xmajorgrids,
xmin=5, xmax=18,
xtick style={color=black},
y grid style={darkgray176},
ymajorgrids,
ymin=1, ymax=10000,
ymode=log,
ytick style={color=black}
]
\addplot [draw=gray, fill=gray, mark=*, only marks]
table{%
x  y
7 1.39
8 1.63
11 1.64
12 935.95
13 1360.0
14 1900.0
15 2590.0
};
\addlegendentry{COR}
\addplot [draw=blue, fill=blue, mark=-, only marks]
table{%
x  y
11 1.4
13 1.41
14 105.83
15 177.87
16 973.97
};
\addlegendentry{EA}
\addplot [draw=black, fill=black, mark=+, only marks]
table{%
x  y
8 1.4
9 185.45
10 194.23
11 622.19
12 710.95
13 1430
14 3240
15 4780
};
\addlegendentry{LA}
\addplot [draw=orange, fill=orange, mark=x, only marks]
table{%
x  y
7 1.4
8 1.41
9 208.38
10 247.32
11 265.42
12 634.12
13 889.03
14 1420
};
\addlegendentry{SA}
\addplot [
  draw=darkviolet1910191,
  fill=darkviolet1910191,
  mark options={rotate=90},
  mark=triangle*,
  only marks
]
table{%
x  y
8 1.4
9 1.41
10 121.34
11 161.57
12 842.8
13 1040
14 1170
15 1280
16 1590
};
\addlegendentry{SN}
\addplot [
  draw=darkturquoise0191191,
  fill=darkturquoise0191191,
  mark options={rotate=270},
  mark=triangle*,
  only marks
]
table{%
x  y
8 1.4
9 140.54
10 162.25
11 412.62
12 825.58
13 1300
14 2050
15 2240
16 2790
};
\addlegendentry{LN}
\addplot [draw=green01270, fill=green01270, mark options={rotate=180}, mark=triangle*, only marks]
table{%
x  y
12 1.4
14 1.41
15 170.85
16 353.53
};
\addlegendentry{EN}
\addplot [draw=red, fill=red, mark=asterisk, only marks]
table{%
x  y
10 1.4
11 1.41
12 106.08
13 107.43
14 122.61
15 127.52
16 339.07
17 1640
};
\addlegendentry{ET}
\end{axis}

\end{tikzpicture}

%% file: plots/uns_champ_time.tex
\begin{tikzpicture}

\definecolor{darkgray176}{RGB}{176,176,176}
\definecolor{darkturquoise0191191}{RGB}{0,191,191}
\definecolor{darkviolet1910191}{RGB}{191,0,191}
\definecolor{green01270}{RGB}{0,127,0}
\definecolor{lightgray204}{RGB}{204,204,204}

\begin{axis}[
legend cell align={left},
legend style={
  fill opacity=0.8,
  draw opacity=1,
  text opacity=1,
  at={(0.03,0.97)},
  anchor=north west,
  draw=lightgray204
},
log basis y={10},
tick align=outside,
tick pos=left,
x grid style={darkgray176},
xlabel={Cumulative instances solved},
xmajorgrids,
xmin=0, xmax=12,
xtick style={color=black},
y grid style={darkgray176},
ylabel={Time taken in sec (log scale)},
ymajorgrids,
ymin=0.1, ymax=100000,
ymode=log,
ytick style={color=black}
]
\addplot [
  draw=darkviolet1910191,
  fill=darkviolet1910191,
  mark options={rotate=90},
  mark=triangle*,
  only marks
]
table{%
x  y
2 0.1
3 3
4 40
5 244
6 372
7 534
8 1608
};
\addlegendentry{SN}
\addplot [
  draw=darkturquoise0191191,
  fill=darkturquoise0191191,
  mark options={rotate=270},
  mark=triangle*,
  only marks
]
table{%
x  y
3 0.1
4 1
5 8
6 698
7 1761
8 4394
9 6114
};
\addlegendentry{LN}
\addplot [draw=green01270, fill=green01270, mark options={rotate=180}, mark=triangle*, only marks]
table{%
x  y
5 0.1
6 20
7 25
8 39
9 599
10 695
};
\addlegendentry{EN}
\addplot [draw=red, fill=red, mark=asterisk, only marks]
table{%
x  y
5 0.1
6 6
7 16
8 17
9 789
10 839
11 8264
};
\addlegendentry{ET}
\end{axis}

\end{tikzpicture}

%% file: plots/sat_champ_time.tex
\begin{tikzpicture}

\definecolor{darkgray176}{RGB}{176,176,176}
\definecolor{darkturquoise0191191}{RGB}{0,191,191}
\definecolor{darkviolet1910191}{RGB}{191,0,191}
\definecolor{green01270}{RGB}{0,127,0}
\definecolor{lightgray204}{RGB}{204,204,204}

\begin{axis}[
legend cell align={left},
legend style={fill opacity=0.8, draw opacity=1, text opacity=1, draw=lightgray204},
log basis y={10},
tick align=outside,
tick pos=left,
x grid style={darkgray176},
xlabel={Cumulative instances solved},
xmajorgrids,
xmin=0, xmax=6,
xtick style={color=black},
y grid style={darkgray176},
ymajorgrids,
ymin=0.1, ymax=100000,
ymode=log,
ytick style={color=black}
]
\addplot [
  draw=darkviolet1910191,
  fill=darkviolet1910191,
  mark options={rotate=90},
  mark=triangle*,
  only marks
]
table{%
x  y
1 3
2 30
3 169
4 10290
};
\addlegendentry{SN}
\addplot [
  draw=darkturquoise0191191,
  fill=darkturquoise0191191,
  mark options={rotate=270},
  mark=triangle*,
  only marks
]
table{%
x  y
1 3
2 37
};
\addlegendentry{LN}
\addplot [draw=green01270, fill=green01270, mark options={rotate=180}, mark=triangle*, only marks]
table{%
x  y
1 1
2 2
3 18
4 148
5 178
};
\addlegendentry{EN}
\addplot [draw=red, fill=red, mark=asterisk, only marks]
table{%
x  y
1 1
2 4
3 12
4 346
5 765
};
\addlegendentry{ET}
\end{axis}

\end{tikzpicture}

%% file: plots/uns_champ_mem.tex
\begin{tikzpicture}

\definecolor{darkgray176}{RGB}{176,176,176}
\definecolor{darkturquoise0191191}{RGB}{0,191,191}
\definecolor{darkviolet1910191}{RGB}{191,0,191}
\definecolor{green01270}{RGB}{0,127,0}
\definecolor{lightgray204}{RGB}{204,204,204}

\begin{axis}[
legend cell align={left},
legend style={fill opacity=0.8, draw opacity=1, text opacity=1, draw=lightgray204},
log basis y={10},
tick align=outside,
tick pos=left,
x grid style={darkgray176},
xlabel={Cumulative instances solved},
xmajorgrids,
xmin=0, xmax=12,
xtick style={color=black},
y grid style={darkgray176},
ylabel={Mem taken in mb (log scale)},
ymajorgrids,
ymin=1, ymax=100000,
ymode=log,
ytick style={color=black}
]
\addplot [
  draw=darkviolet1910191,
  fill=darkviolet1910191,
  mark options={rotate=90},
  mark=triangle*,
  only marks
]
table{%
x  y
1 1.4
3 1.41
4 125.11
5 482.43
6 622.41
7 721.55
8 2920
};
\addlegendentry{SN}
\addplot [
  draw=darkturquoise0191191,
  fill=darkturquoise0191191,
  mark options={rotate=270},
  mark=triangle*,
  only marks
]
table{%
x  y
4 1.4
5 1.41
6 773.8
7 1280
8 6280
9 25850
};
\addlegendentry{LN}
\addplot [draw=green01270, fill=green01270, mark options={rotate=180}, mark=triangle*, only marks]
table{%
x  y
6 1.4
7 1.41
8 167.35
9 749.64
10 777.24
};
\addlegendentry{EN}
\addplot [draw=red, fill=red, mark=asterisk, only marks]
table{%
x  y
8 1.4
9 706.38
10 746.61
11 2960
};
\addlegendentry{ET}
\end{axis}

\end{tikzpicture}

%% file: plots/sat_champ_mem.tex
\begin{tikzpicture}

\definecolor{darkgray176}{RGB}{176,176,176}
\definecolor{darkturquoise0191191}{RGB}{0,191,191}
\definecolor{darkviolet1910191}{RGB}{191,0,191}
\definecolor{green01270}{RGB}{0,127,0}
\definecolor{lightgray204}{RGB}{204,204,204}

\begin{axis}[
legend cell align={left},
legend style={fill opacity=0.8, draw opacity=1, text opacity=1, draw=lightgray204},
log basis y={10},
tick align=outside,
tick pos=left,
x grid style={darkgray176},
xlabel={Cumulative instances solved},
xmajorgrids,
xmin=0, xmax=6,
xtick style={color=black},
y grid style={darkgray176},
ymajorgrids,
ymin=1, ymax=100000,
ymode=log,
ytick style={color=black}
]
\addplot [
  draw=darkviolet1910191,
  fill=darkviolet1910191,
  mark options={rotate=90},
  mark=triangle*,
  only marks
]
table{%
x  y
1 1.4
2 147.25
3 283.82
4 3390
};
\addlegendentry{SN}
\addplot [
  draw=darkturquoise0191191,
  fill=darkturquoise0191191,
  mark options={rotate=270},
  mark=triangle*,
  only marks
]
table{%
x  y
1 1.4
2 255.45
};
\addlegendentry{LN}
\addplot [draw=green01270, fill=green01270, mark options={rotate=180}, mark=triangle*, only marks]
table{%
x  y
2 1.4
3 1.41
4 471.33
5 537.52
};
\addlegendentry{EN}
\addplot [draw=red, fill=red, mark=asterisk, only marks]
table{%
x  y
3 1.4
4 619.13
5 791.27
};
\addlegendentry{ET}
\end{axis}

\end{tikzpicture}